\documentclass{ametsocV6.1} 



\usepackage{amsmath,amsfonts,amssymb,bm}
\usepackage{mathptmx}
\usepackage{newtxtext}
\usepackage{newtxmath}
\usepackage{float}
\usepackage[normalem]{ulem}
\usepackage{xr}

\usepackage{authblk}

\makeatletter
\def\fps@figure{H}
\makeatother

\title{Evidential Deep Learning: Enhancing Predictive Uncertainty Estimation for Earth System Science Applications}

\authors{
John S. Schreck\aff{a}\correspondingauthor{John S. Schreck, schreck@ucar.edu},
David John Gagne II\aff{a},
Charlie Becker\aff{a},
William E. Chapman\aff{a},
Kim Elmore\aff{b},
Da Fan\aff{h},
Gabrielle Gantos\aff{a},
Eliot Kim\aff{a},
Dhamma Kimpara\aff{c},
Thomas Martin\aff{d},
Maria J. Molina\aff{e,a},
Vanessa M. Przybylo\aff{f},
Jacob Radford\aff{g},
Belen Saavedra\aff{a},
Justin Willson\aff{a},
Christopher Wirz\aff{a}
}

\affiliation{
\aff{a}{NSF National Center for Atmospheric Research, Boulder, CO, USA}\\
\aff{b}{Cooperative Institute for Severe and High-Impact Weather Research and Operations, National Severe Storms Laboratory, Norman, OK, USA}\\
\aff{c}{Department of Computer Science, University of Colorado, Boulder, CO, USA}\\
\aff{d}{University Corporation for Atmospheric Research, Unidata, Boulder, CO, USA}\\
\aff{e}{Department of Atmospheric and Oceanic Science, University of Maryland, College Park, MD, USA}\\
\aff{f}{University at Albany, State University of New York, Albany, New York, USA}\\
\aff{g}{Cooperative Institute for Research in the Atmosphere, Colorado State University, Fort Collins, CO, USA}\\
\aff{h}{Department of Meteorology and Atmospheric Science, The Pennsylvania State University, University Park, PA, USA}
}

\abstract{Robust quantification of predictive uncertainty is critical for understanding factors that drive weather and climate outcomes. Ensembles provide predictive uncertainty estimates and can be decomposed physically, but both physics and machine learning ensembles are computationally expensive. Parametric deep learning can estimate uncertainty with one model by predicting the parameters of a probability distribution but do not account for epistemic uncertainty. Evidential deep learning, a technique that extends parametric deep learning to higher-order distributions, can account for both aleatoric and epistemic uncertainty with one model. This study compares the uncertainty derived from evidential neural networks to those obtained from ensembles. Through applications of classification of winter precipitation type and regression of surface layer fluxes, we show evidential deep learning models attaining predictive accuracy rivaling standard methods, while robustly quantifying both sources of uncertainty.  We evaluate the uncertainty in terms of how well the predictions are calibrated and how well the uncertainty correlates with prediction error. Analyses of uncertainty in the context of the inputs reveal sensitivities to underlying meteorological processes, facilitating interpretation of the models. The conceptual simplicity, interpretability, and computational efficiency of evidential neural networks make them highly extensible, offering a promising approach for reliable and practical uncertainty quantification in Earth system science modeling. In order to encourage broader adoption of evidential deep learning in Earth System Science, we have developed a new Python package, MILES-GUESS (\url{https://github.com/ai2es/miles-guess}), that enables users to train and evaluate both evidential and ensemble deep learning.}

\begin{document}

\maketitle

\noindent\textbf{Significance Statement}

This study demonstrates a new technique, evidential deep learning, for robust uncertainty quantification in modeling the Earth system. The method integrates probabilistic principles into deep neural networks, enabling the estimation of both aleatoric uncertainty from noisy data and epistemic uncertainty from model limitations using a single model. Our analyses reveal how decomposing these uncertainties provides valuable insights into reliability, accuracy, and model shortcomings. We show the approach can rival standard methods in classification and regression tasks within atmospheric science while offering practical advantages such as computational efficiency. With further advances, evidential networks have the potential to enhance risk assessment and decision-making across meteorology by improving uncertainty characterization, a longstanding challenge. This work establishes a strong foundation and motivation for broader adoption of evidential learning where properly quantifying uncertainties is critical yet lacking.

\section{Introduction} 

Uncertainty is an inherent aspect of any prediction \citep{abdar2021review}, yet it is often overlooked or not communicated effectively alongside the prediction itself \citep{gneiting2007probabilistic}. The ability to provide well-calibrated and robust predictive uncertainty estimates can be highly valuable, allowing users to understand the reliability of predictions and make more informed decisions based on them \citep[e.g.,][]{Nadav_Greenberg_2009, kendall2017}. For model developers, accurate predictive uncertainty estimates can help identify challenging cases and determine when a model may be operating outside its training domain \citep{kendall2017, karpatne2018}. Additionally, by connecting uncertainty estimates with other analysis tools, researchers can gain insights into the input sensitivities that influence uncertainty levels, allowing a better understanding of the factors that drive these uncertainties \citep{herman2018money, liu2020simple}. In the realm of machine learning, various approaches have emerged for quantifying total predictive uncertainty. These include well-established methods such as bagging \citep{breiman1996bagging}, Gaussian processes \citep{rasmussen2006gaussian}, quantile regression \citep{koenker2005quantile}, as well as newer conformal methods \citep{romano2019conformalized, stankeviciute2021conformal, angelopoulos2021gentle}. However, it is essential to recognize that existing methods often come with inherent limitations. Conventional techniques, such as bagging, may encounter difficulties in capturing the full spectrum of uncertainty, particularly when dealing with intricate, multimodal distributions. Moreover, many established methods often lack the ability to decompose predictive uncertainty into its underlying components, hindering a deeper understanding of uncertainties. Additionally, handling custom probability distributions can be challenging for some of these methods, limiting their adaptability to specific problem domains.  

Traditionally, uncertainty quantification (UQ) within the Earth System Science community has been pursued through the development and enhancement of physics-based numerical models, which generate probabilistic forecasts using ensembles of deterministic forecasts that vary in initial conditions, boundary conditions, or model specifications \citep{Leith1974}. One notable advantage of these methods is their strong foundation in the true physics of atmospheric/oceanic motion. However, deterministic numerical model ensembles come with considerable computational costs and often lack proper uncertainty calibration \citep{vannitsem2018statistical}. To combat the computational expense and lack of calibration, UQ has been attempted through statistical linking functions. Two prominent techniques are ensemble model output statistics \citep{gneiting2005calibrated}, in which a parametric distribution is prescribed and fit, and Bayesian model averaging \citep{raftery2005using}, where the UQ takes the form of a weighted mixture distribution.

The use of modern Machine Learning (ML) for Earth system UQ has been the focus of much recent research \citep{mcgovern2017using, haynes2023creating}, especially within the forecast post-processing community \citep{haupt2021towards, schulz2022machine, vannitsem2020statistical}. Popular techniques include parametric fitting \citep{ghazvinian2021novel, rasp2018neural, chapman2022probabilistic, guillaumin2021stochastic, barnes2021controlled, foster2021probabilistic, Delaunay2022, gordon2022incorporating}, quantile-based probabilities transformed to full predictive distributions \citep{scheuerer2020using}, or creating direct approximations of the quantile function via regression based on Bernstein polynomials \citep{bremnes2020ensemble}.

Many users of uncertainty rely on a single metric, such as probability or ensemble spread, but decomposing uncertainty into different components provides valuable insights into its sources and nature \citep{kendall2017}, and can help validate how well a model captures the different uncertainty sources. In statistics, the law of total variance decomposes uncertainty into \textit{aleatoric}, arising from inherent data randomness, and \textit{epistemic}, arising from model limitations and insufficient training data \citep{kendall2017}. High aleatoric uncertainty indicates the lack of a clear relationship between the model inputs and the target and can only be reduced with the addition of more informative input variables \citep{herman2018money}. High epistemic uncertainty can be reduced by accumulating more data in sparse areas of the input space or by reducing the complexity or flexibility of the model. Such distinction aids in assessing model learning capacity, generalization, and guiding hyperparameter optimization \citep{karpatne2018}. Advancing this field requires techniques that efficiently decompose uncertainty while achieving general predictive reliability.

While parametric probabilistic machine learning models, such as those that predict the parameters of a categorical or Gaussian probability distribution, can express aleatoric uncertainty \citep{nix1994estimating}, they do not account for epistemic uncertainty \citep{amini2020}. The predicted variance only accounts for data variance, not model variance. Ensembles of deterministic ML models \citep{lakshminarayanan2016simple} and sampling methods, such as Monte Carlo dropout \citep{srivastava2014dropout, gal2016dropout} approximate epistemic uncertainty by deriving their spread from model perturbations, but if their predictions are single labels or values, then they are not accounting for spread in the data. On the other hand, Bayesian neural networks \citep{neal2012bayesian}, which treat every weight as a random variable, can accurately estimate both aleatoric and epistemic uncertainty but are computationally demanding and challenging to implement for complex architectures. Ensembles of parametric probabilistic models can also be used to approximate aleatoric and epistemic uncertainty \citep{Delaunay2022} but only with a sufficiently large ensemble size \citep{shaker2020aleatoric}.

The recent rise of evidential neural networks \citep{sensoy2018evidential, amini2020, ulmer2023} offers a promising solution that strikes a balance between efficiency and accuracy while providing an effective approach to estimate both sources of uncertainty. Evidential neural networks use a single deterministic neural network while modifying the prediction task to estimate the parameters of a higher-order evidential distribution, which draws relevance from Bayesian data analysis principles \citep{gelman2013bayesian}. This distribution treats the parameters of the target distribution as random variables and models them with an assumed prior distribution \citep{sensoy2018evidential, amini2020}. For multinomial (categorical) and Gaussian distributions, analytical formulations of conjugate prior distributions, such as the Dirichlet and Normal Inverse Gamma \citep{sensoy2018evidential, amini2020}, enable the construction of exact loss functions for neural network training. These loss functions consist of a negative log-likelihood component to maximize the fit to the data and a regularizer term to minimize evidence allocated to incorrect predictions and inflate the conditional uncertainty \citep{sensoy2018evidential}.

In this work, we introduce the concept of evidential deep learning to the Earth system science community. Evidential deep learning represents a relatively recent machine learning technique known for its ability to provide predictive uncertainty estimates with practical advantages, as discussed by \cite{sensoy2020uncertainty}. Our primary focus is on the application of evidential neural networks within the weather and climate domain, where the accurate estimation of uncertainty holds significance in decision-making, as highlighted in previous studies \citep{shepherd2009geoengineering, bauer2015quiet}. To assess the utility of evidential neural networks, we employ them in both classification and regression tasks. In the classification domain, we showcase an evidential neural network trained to predict winter precipitation type based on simulated atmospheric soundings. In regression tasks, our evaluation centers on assessing the model's performance in estimating surface energy fluxes using observed meteorological variables \citep{mccandless2022machine}.

In addition to practical applications, we have defined three key objectives: (1) To quantitatively assess and compare the predictive capabilities of evidential versus deterministic neural networks using essential metrics such as Brier Skill Score (BSS) and Root Mean Square Error (RMSE). (2) To evaluate the calibration of predicted uncertainties derived from evidential models and ensembles through various analysis techniques. (3) To explore calibration tuning approaches tailored to evidential networks, including parameter adjustments such as a loss regularization weight for regression and the fine-tuning of the dropout rate for Monte Carlo (MC) ensembles. These objectives aim to provide a balanced understanding of the potential of evidential architectures in generating uncertainty estimates while maintaining accuracy, and emphasize the significance of uncertainties in assessing the limitations of ML models.

\section{Methods}

\subsection{Example Problems of Interest}

\begin{figure}[t]
    \centering
    \includegraphics[width=\columnwidth]{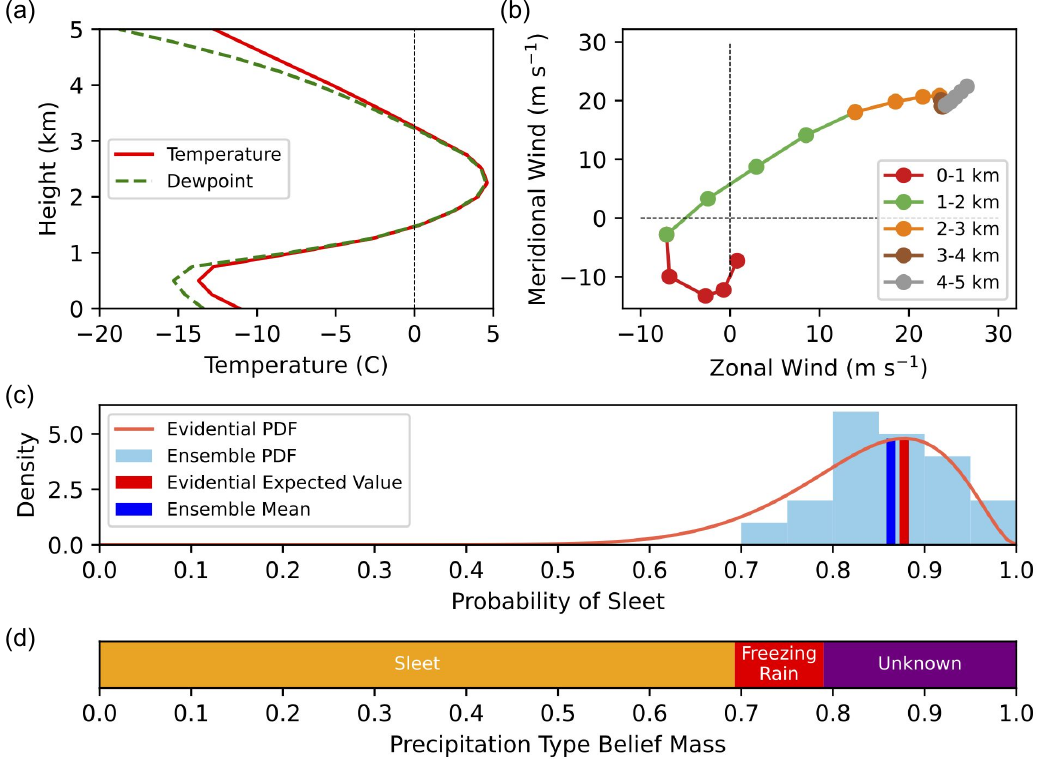}
    \caption{(a-b) Example of a precipitation type sounding with temperature, dewpoint, and wind profiles along with (c-d) different predictive uncertainty representations.}
    \label{fig:evidential-1}
\end{figure}

In the realm of supervised learning, a neural network is employed to predict the most likely label or outcome based on a given set of input predictor variables. Figure~\ref{fig:evidential-1} provides an illustration of model inputs and outputs for a classification problem, based on one of the example problems explored in this paper. The inputs are shown in Figure~\ref{fig:evidential-1}(a) and (b) and include temperature ($T$; $^\circ$C), dewpoint temperature ($T_\text{dew}$; $^\circ$C), and zonal ($U$; m s$^{-1}$) and meridional ($V$; m s$^{-1}$) wind, at equally spaced heights in the atmosphere. The output is a winter precipitation-type label derived from a set of precipitation-type probabilities: $\mathbf{p}$ = $(p_{rain}, p_{snow}, p_{sleet}, p_{frzrain})$ (Figure~\ref{fig:evidential-1}(c)). 

The contrast between the two modeling approaches investigated here (ensemble versus evidential) when predicting the `sleet' label is illustrated in Figure~\ref{fig:evidential-1}(c). In the case of the evidential model, the predictions are represented by a smooth curve, offering a continuous and comprehensive quantification of possible outcomes. Conversely, an ensemble approach such as k-fold cross validation or Monte Carlo sampling, provides discrete point estimates, as seen in the figure, which are obtained by collecting multiple model predictions. It's important to note that these point estimates require a full ensemble to approximate the continuous probabilistic insights provided directly by the evidential model. This distinction highlights the effectiveness of the evidential model in capturing the full spectrum of uncertainty for the full set of labels, as opposed to the more discrete and ensemble-based representation of uncertainty.

The data used for winter precipitation-type modeling consists of two main sources. The target data source is the NOAA National Severe Storms Laboratory Meteorological Phenomena Identification Near the Ground, or mPING \citep{elmore2014mping}. mPING is a crowd-sourced, GPS-location-linked, weather reporting project that allows anyone to report precipitation type and other weather hazards anonymously across the globe. For this project, we have selected only snow, rain, freezing rain, and sleet/ice pellet reports and have narrowed the data window to be between 1 January 2015 and 30 June 2022. Data after 1 July 2020 were withheld for testing and training and validation data was split randomly after being grouped by day. The value of these reports come from being able to gather information from any populated location rather than at fixed weather stations, but the reports contain biases and uncertainties related to users' abilities to discern different precipitation types from each other.

The input data are simulated vertical atmospheric profiles from the NOAA Rapid Refresh (RAP) numerical weather prediction model analysis \citep{Benjamin2016-mj}.  The RAP has a 13 km horizontal resolution on a Lambert Conformal grid over North America. RAP temperature, relative humidity, wind, and pressure are interpolated from pressure levels ranging every 25 hPa from 1000 to 100 hPa to height levels every 250 m from 0 to 5 km above ground level. Dewpoint is derived from relative humidity, temperature and pressure using the metpy package \citep{May2022-fy}. Surface temperature, dewpoint, and wind are extracted from the RAP 2 m temperature and dewpoint and 10 m wind components. mPING observations are grouped by hour within the bounds of each RAP grid cell. The most common precipitation type is used as the target for the machine learning models. 

Within the matched mPING and RAP data, a small percentage of cases contained reports that were physically inconsistent with the RAP profile. These cases were found after initial neural networks trained on the data reported nonzero probabilities of frozen precipitation even when near-surface temperatures were well above 0 $^{\circ}$C. Further investigation revealed many profiles labeled as sleet and freezing rain but lacking a profile where the wet bulb temperature goes above 0 $^{\circ}$C before returning to below freezing. Some reports were labeled with a frozen type but did not have any freezing layers near the surface or near surface moisture. When matched with radar, some of these reports were not colocated with any precipitation on radar. To address these inconsistencies, we filtered examples where there was a frozen precipitation type (snow, freezing rain, or sleet) reported but the surface wet-bulb temperature exceeded 3 $^{\circ}$C or if rain was reported and the surface wet-bulb temperatures was less than -3 $^{\circ}$C.  We tested other wet-bulb temperature thresholds up to 10 $^{\circ}$C and found that 3 $^{\circ}$C filtered most of the inconsistent reports without removing too many plausible reports.

\begin{figure}[t!]
    \centering
    \includegraphics[width=\columnwidth]{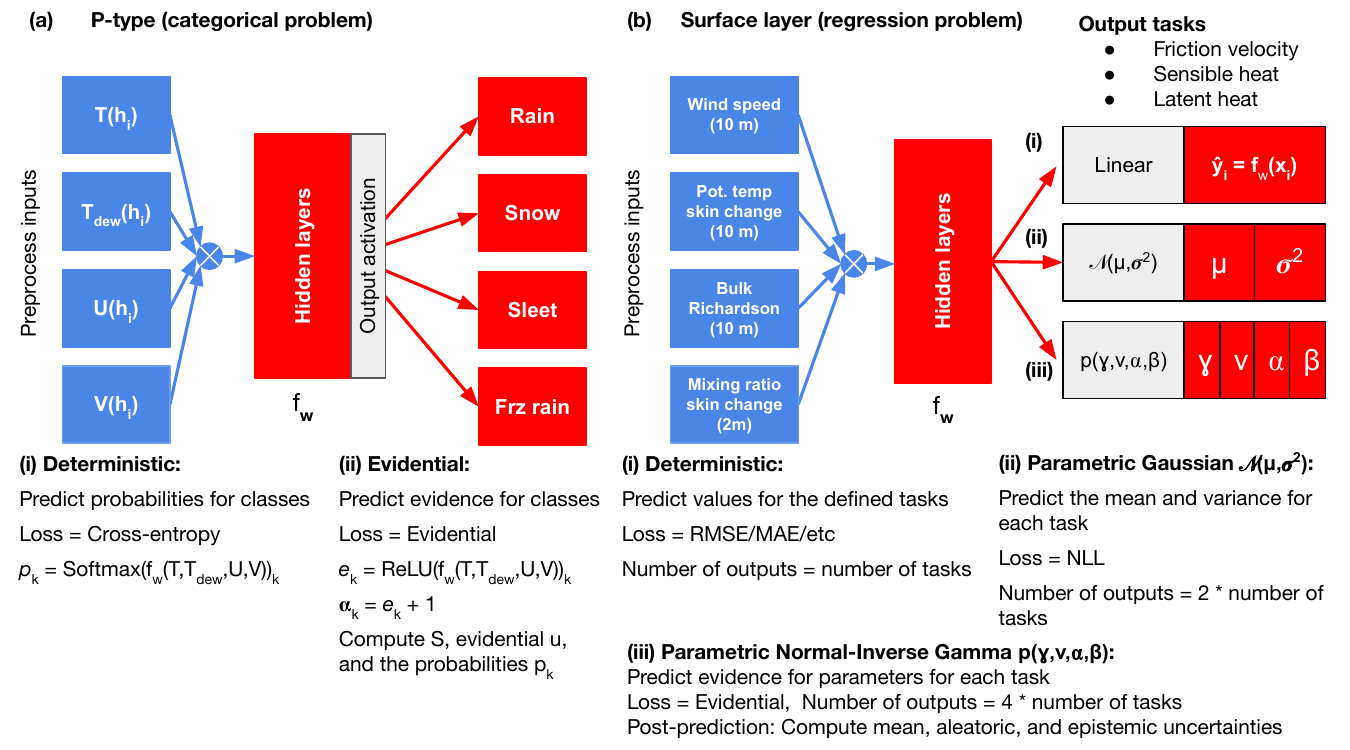}
    \caption{(a) Deterministic and evidential MLP architectures for predicting class probabilities in the precipitation-type categorical dataset. (b) Architectures for predicting parameters in the surface-layer regression dataset, including Gaussian ($\mu$, $\sigma^2$) and Normal-Inverse Gaussian ($\gamma$, $\nu$, $\alpha$, $\beta$) distributions. In both architecture diagrams, the neural network is represented by $\text{f}_\textbf{w}$ where $\textbf{w}$ are trainable parameters.}
    \label{fig:models}
\end{figure}

Figure~\ref{fig:models}(a) illustrates a multi-layer perceptron (MLP) neural network architecture tasked with predicting precipitation-type class probabilities that we will investigate here. In a similar vein, as illustrated in Figure~\ref{fig:models}(b), we employ another MLP architecture, but this time for a regression problem. The model predicts energy fluxes from Earth's surface given near-surface atmospheric conditions \citep{mccandless2022machine, Munoz-Esparza2022-rm}. The ultimate goal of this problem is to train a machine learning parameterization of the surface layer from observations that could be used to replace existing parameterizations in weather and climate models. In this example, the model takes as input four quantities: wind speed (m s$^{-1}$) at 10 m above ground level (AGL), the change in potential temperature (K s$^{-1}$) between the surface and 10 m AGL, the Bulk Richardson number (unitless) valid at 10 m AGL, and the change in water vapor mixing ratio (g kg$^{-1}$ m$^{-1}$) between the surface and 10 m AGL. The illustrated architecture predicts up to three outputs: friction velocity (m s$^{-1}$), kinematic sensible heat flux (kg m s$^{-1}$), and kinematic latent heat flux (kg m s$^{-1}$). 

The dataset used in the surface layer energy flux problem (SL) provides long-term observations from the Royal Netherlands Meteorological Institute (KNMI) Cabauw Experimental Site for Atmospheric Research located in Cabauw, Netherlands \citep{Bosveld2020-kg}. The site has been utilized for validating surface-layer parameterizations and land-surface models since 1972. The area surrounding the site consists of flat agricultural land with minimal elevation changes and low brush coverage.

We used the Cabauw data gather from 2003-2017 for training and testing the ML models. The dataset includes various meteorological measurements noted earlier that are used as inputs to the model as well as model training targets \citep{mccandless2022machine, munoz2022application}. The outgoing longwave radiation, converted to skin temperature, is derived from a device located approximately 200 meters from the flux tower. The data were grouped by day and then split randomly into training, validation, and testing sets. The same random seed was used for splitting to ensure that all ML models used the same testing set. \citet{mccandless2022machine} recently showed that the Cabauw dataset could be used to train effectively several machine learning models including a MLP like the one illustrated in Figure~\ref{fig:models}(a) with good performance. 

\subsection{Classification Challenges}
Once trained, the ML models are used to make predictions for a series of events. We define an event as the space and time extent over which a single prediction is valid, such as from 1800 to 1900 UTC on 2 February 2023 in the area within 13 km of the center of Boulder, Colorado. In classification problems, ML models output a number associated with each of $K$ known outcomes for an event, given a vector of inputs $\mathbf{x}$. If the numbers each range between 0 and 1 and sum to 1, then the numbers behave as probabilities and thus represent the parameters of a categorical distribution describing a constant probability of outcome occurrence during the event. 
\begin{equation}
p(\hat{y}=y_k|\mathbf{x})=p_k(\mathbf{x})
\end{equation}
Although predictions cover the volume of an event, the observations that verify the predictions generally occur only at a single point in space and time within that event. In the winter precipitation type case, the observed outcome is a series of crowd-sourced reports occurring within a given area and time frame that may vary from the most likely outcome due to a mix of subscale meteorological and societal factors. To model the potential variability in observed outcomes for an event, we can use the multinomial distribution, which generalizes the categorical distribution to the outcomes of $m$ repeated observations under the same event conditions. Mathematically, we express the multinomial probability for a set of $m$ observations with a fixed number for each outcome $\mathbf{y}=\{y_k,...,y_K\}$ where $\sum_{k=1}^K y_i=m$, given a fixed set of probabilities for each outcome $\mathbf{p}=\{p_k,...,p_K\}$ as
\begin{equation}
\label{eq:multinomial}
p(\mathbf{y}| m, \mathbf{p}, K) = \frac{{m!}}{{\prod\limits_{k=1}^K y_k!}} \prod_{k=1}^K p_k^{y_k}.
\end{equation}

To train the ML model to estimate $\mathbf{p}$ for a given event, we construct a training dataset $\mathcal{D}$ = $\{\mathbf{x}_n, \mathbf{y}_n\}_{n=1}^{N}$ that we then input into the ML model repeatedly in order to adjust the parameters (weights) $\mathbf{w}$ of the ML model to maximize their likelihood given the training data. This statistical optimization process, called maximum likelihood estimation, involves the model ingesting the inputs and predicting outputs y to minimize the negative log-likelihood loss function 
\begin{equation}
\mathcal{L}(\mathbf{w}| \mathcal{D}) = - \frac{1}{N} \sum_{n=1}^N \sum_{k=1}^K y_{n,k} \ln(\hat{y}_{n,k}(\mathbf{x}_n))
\end{equation}
where $N$ is the number of data points, $K$ is the number of classes, $\mathbf{w}$ are the model parameters, $y_{n,k}$ is a binary indicator for whether the n-th sample belongs to class $k$, and $\hat{y}_{n,k}(\mathbf{x}_n)$ is the predicted probability of the n-th sample belonging to class $k$. Maximum likelihood estimation assumes that there we are seeking one fixed set of $\mathbf{w}$ that best explains our data due to its origins in frequentist statistics \citep{mackay2003information}. Thus, these predicted probabilities only account for aleatoric uncertainty from the variation in the outcome for each input.

\subsection{Model Ensembles}

Additional model weights could also plausibly explain the data and these plausible model weights could be found by repeating the maximum likelihood estimation process with different random initializations or by resampling the training data. Both approaches can produce ensembles of probabilistic ML models \citep{lakshminarayanan2016simple, gal2016dropout, dietterich2000ensemble}. Another approach is the Bayesian neural network \citep{neal2012bayesian}, which treats the weights of the neural network as random variables that are each represented with a parametric probability distribution. Both approaches allow for the estimation of both aleatoric and epistemic uncertainty, extending the model's ability to capture and express uncertainty in predictions. Ensembles are straightforward to train but often require a large number of members or many repeated samples to produce robust uncertainty estimates. Bayesian neural networks require more weights per model compared with standard MLPs and are less likely to converge to results that both perform well deterministically and produce robust uncertainty estimates. 

To generate ensembles, we will consider three methods: cross-validation splitting, deep ensembling \citep{lakshminarayanan2016simple}, and Monte Carlo dropout \citep{gal2016dropout}. Cross-validation is a standard approach for training and evaluating the performance of neural models. It involves dividing the dataset into multiple subsets or folds, using a fold as a validation set while the remaining folds form the training set. The ensemble model is then both trained and evaluated multiple times, with each fold serving as the validation set for a different model. Ensembles of k-fold=20 were used. 

Deep ensembling \citep{lakshminarayanan2016simple} extends the concept of ensemble methods to deep learning architectures. Instead of training a single neural network, multiple instances of the same network are independently trained. Each network explores different regions of the parameter space due to random initialization or training dynamics. To create the ensemble, we employ different weight parameterizations of the neural networks using fixed seeds, ensuring that all models in the ensemble are trained on the same data split. Similarly, ensembles of size 20 were used. 

Finally, Monte Carlo dropout \citep{gal2016dropout} involves randomly deactivating neurons during model training using a dropout rate. During inference, the model is used to obtain $N _{MC}$ predictions while keeping dropout enabled. We used $N _{MC} = 100$ in all cases to create the ensembles of predictions. Note that all of these ensemble generation methods are applicable to both categorical and regression problems.

\subsection{Evidential Classification}

The recent evidential deep learning approaches find a compromise between ensemble methods and Bayesian neural networks by using fixed weights for the neural network but predict instead the parameters for a higher-order posterior distribution that describes the space of possible lower-order distributions that could only be partially sampled by ensembles \citep{sensoy2018evidential, amini2020, meinert2021, oh2022improving, meinert2023unreasonable}. In Bayesian inference, prior information is combined with the observed data using Bayes' theorem to update our beliefs and obtain the posterior distribution. Recall that Bayes' theorem states that the posterior probability of a parameter given the data is proportional to the product of the prior probability and the likelihood of the data given the parameter: Posterior $\sim$ Prior $\cdot$ Likelihood. By including a prior distribution in our inference process, we can define the space of possible models and thus include the epistemic uncertainty. However, in order to make the inference problem tractable, we must pick a prior distribution that can be analytically related to our posterior distribution.

Given these constraints, the best choice of prior for our multinomial/categorical target distribution is the Dirichlet distribution \citep{murphy2007conjugate, hoffman2013stochastic}. It is described by hyperparameters $\boldsymbol{\alpha}= (\alpha_1, \alpha_2, ..., \alpha_K)$, the probability density function has the following form
\begin{equation}
f(\mathbf{p}|\boldsymbol{\alpha}) = 
\begin{cases}
    \frac{1}{B(\boldsymbol{\alpha})} \prod\limits_{k=1}^{K} p_k^{\alpha_k - 1} & \text{for } \textbf{p} \in S_K, B(\boldsymbol{\alpha})=\frac{\prod\limits_{k=1}^{K}\Gamma(\alpha_k)}{\Gamma(\sum\limits_{k=1}^{K}a_k)}\\
    0              & \text{otherwise},
\end{cases}
\end{equation}
where $S_K$ is the unit simplex defined as $S_K$ = $\{\mathbf{p} | \sum_{k=1}^K p_k = 1~\text{and}~0\le p_1 ,...,p_K\le 1 \}$, and $B(\boldsymbol{\alpha})$ is the K-dimensional multinomial beta function \citep{kotz2004continuous}. Note the similar form to Equation \ref{eq:multinomial}. The expected probability for the $k$th outcome is the mean of the Dirichlet posterior distribution
\begin{equation}
\mathbb{E}[p_k] = \hat{p}_k = \frac{\alpha_k}{S}.
\end{equation}
where $S \equiv \sum_{k=1}^K \alpha_k$. The Dirichlet is chosen because it is the conjugate prior to the multinomial, meaning that the posterior when the Dirichlet prior is combined with a multinomial likelihood is also a Dirichlet. This drastically simplifies the updating process during Bayesian inference because each posterior $\alpha_k$ can be computed by summing the prior $\alpha_k$ with each observed or predicted $y_k$. In this context, each $\alpha_k$ is a pseudocount of how many outcomes of a certain type are expected to occur during the event. 

\citet{sensoy2018evidential} also motivated the use of the Dirichlet posterior distribution through the Dempster-Shafer Theory of Evidence (DST), which is an extension of Bayesian statistics \citep{dempster1968generalization} to decision-making with uncertain evidence. Evidence in the DST classification context is represented as a non-negative number $e_k$ with higher values indicating stronger evidence for a particular outcome. The uninformed evidence prior is that each outcome has evidence of 1, so summing the prior evidence with how much evidence is observed for each outcome from data or a predictive model, we receive a posterior $\alpha_k$ = $e_k + 1$ and the total amount of evidence $S = \sum_{k=1}^K (e_k + 1)$. The $\alpha$ values can then be plugged into a Dirichlet distribution to derive probabilities for each outcome.

DST is extended for decision-making in classification problems through Subjective Logic \citep{jsang2018subjective}. Subjective Logic, as a formal framework for modeling uncertainty and subjective beliefs, utilizes belief masses to quantify belief strength and enables nuanced representation of subjective confidence levels. Within this framework, a belief mass $b_k$ is the amount of evidence assigned to a particular outcome, and uncertainty mass is the amount of evidence not allocated to any outcome, 
\begin{eqnarray}
b_k &= \frac{e_k}{S}, \\ 
u &= \frac{K}{S}.
\end{eqnarray}
$K$ is usually defined as the number of outcomes, assuming no prior evidence for any outcome. Belief masses look similar to probabilities but only have to sum to 1 when $u$ is included:
\begin{equation}
u + \sum_{k=1}^K b_k = 1.
\end{equation}
The color bar in Figure~\ref{fig:evidential-1}(d) illustrates the belief masses plus $u$ for a precipitation-type evidential model. 

By incorporating prior knowledge through base rates, Subjective Logic contextualizes beliefs within a broader information framework. In the integration of DST and Subjective Logic, the Dirichlet distribution provides a natural framework for representing belief masses, which are used to quantify the strength of subjective beliefs. The Dirichlet distribution also enables flexible modeling of uncertainty by capturing the covariance among different belief masses.

Figure~\ref{fig:models}(a)(ii) shows an MLP architecture employing the `evidential' categorical model. The parametric neural network parameterized by $\mathbf{w}$, and given an input sample $\mathbf{x}_n$, now predicts the evidence vector, $e_k$, represented by $f(\mathbf{x}_n | \mathbf{w})$, hence the name assignment `evidential MLP'. Compared to a deterministic classifier, the only architectural difference is the output activation function, which is taken to be ReLU following \cite{sensoy2018evidential} rather than softmax, to filter non-negative evidence. Accordingly, the Dirichlet distribution corresponding with this evidence has parameters $\alpha_k$ = $f(\mathbf{x}_n | \mathbf{w}) + 1$. The predicted expected probabilities for the classes are then computed as $\alpha_k / S$. 

\citet{sensoy2018evidential} explored using one loss based on Type-II Maximum Likelihood, and \citet{malinin2018predictive} focused on computing the Bayesian risk with respect to the class predictor. Here we only present one of the risk losses as the authors noted the superior performance. Specifically, the parametric neural network employs the following loss during training
\begin{eqnarray}
\label{nig_loss}
\mathcal{L}_n(\textbf{w}) &= \int ||\mathbf{y}_n - \mathbf{p}_n||^2 \frac{1}{B(\boldsymbol{\alpha_n})} \prod\limits_{k=1}^K p_{n,k}^{\alpha_{n,k}-1}d\mathbf{p}_n \\ 
&= \sum\limits_{k=1}^K \left( y_{n,k} - \hat{p}_{n,k} \right)^2 + \frac{\hat{p}_{n,k}(1-\hat{p}_{n,k})}{S + 1}.
\end{eqnarray}
\citet{sensoy2018evidential} prove three propositions to explain how Eq.~\ref{nig_loss} optimizes the models: (1) The proposed loss function in the neural network prioritizes data fit by minimizing the variance of the generated Dirichlet experiment for each sample in the training set. (2) It decreases the prediction error when new evidence is added to the Dirichlet parameters and increases when evidence is removed, ensuring effective learning from new information. (3) Additionally, the loss function exhibits learned loss attenuation, reducing prediction error by removing evidence from the largest Dirichlet parameter for a sample, except when the class label is the correct one. 

By combining these propositions, it is concluded that the neural network with the proposed loss function generates more evidence for the correct class labels and avoids misclassification by removing excessive misleading evidence. The loss function also aims to decrease the variance of predictions on the training set by increasing evidence, but only if it leads to a better data fit. The loss over a batch of training samples is computed by summing the loss for each individual sample. The model learns to generate evidence for specific class labels based on patterns in the data, and it adjusts the generated evidence based on backpropagation when counter-examples are observed. However, limited counter-examples may result in an increased overall loss when reducing the evidence magnitude, leading to some misleading evidence for incorrect labels.

To address this limitation, a Kullback-Leibler (KL) divergence term is incorporated into the loss function
\begin{equation}
\label{total_loss}
    \mathcal{L}(\textbf{w}) = \sum_{i=1}^N \mathcal{L}_n(\textbf{w}) + \nu_t \sum_{n=1}^N KL\left[ D(\mathbf{p}_n | \widetilde{\boldsymbol{\alpha}_n}) || D(\mathbf{p}_n|\mathbf{1})\right],
\end{equation}
where $\nu_t$ is an annealing coefficient that depends on the epoch number $t$, $D(\mathbf{p}_n|\mathbf{1})$ represents a uniform Dirichlet distribution, $\mathbf{1}$ is a vector of ones, and $\widetilde{\boldsymbol{\alpha}_n} = \mathbf{y}_i + (1-\mathbf{y}_i) \odot \boldsymbol{\alpha}_n$ represents the Dirichlet parameter after the misleading evidence has been removed for sample $n$, where $\odot$ represents the element-wise (Hadamard) product.

This term regularizes the predictive distribution and penalizes divergences from a state of uncertainty (``I don't know'') that do not contribute to data fit. The loss function with this regularization term is used to guide the model's behavior and achieve the desired total evidence shrinkage to zero for samples that cannot be correctly classified. The KL divergence term can be reduced to
\begin{eqnarray}
    KL\left[ D(\mathbf{p}_n| \widetilde{\mathbf{\alpha}_n}) || D(\mathbf{p}_n| \mathbf{1}) \right] &= \log\left( \frac{\Gamma\left(\sum_{k=1}^K \widetilde{\alpha}_{n,k}\right)}{\Gamma(K) \prod_{k=1}^K \Gamma(\widetilde{\alpha}_{n,k})}\right) \nonumber \\
    &+ \sum\limits_{k=1}^K \left(\widetilde{\alpha}_{n,k}-1 \right) \left[ \psi(\widetilde{\alpha}_{n,k}) - \psi\left( \sum\limits_{j=1}^K \widetilde{\alpha}_{n,j} \right) \right],
\end{eqnarray}
where $\psi$ represents the digamma function, $\Gamma$ is the Gamma function. By incrementally increasing the impact of the KL divergence in the loss function via the annealing coefficient, the neural network is allowed the opportunity to explore different parameter configurations. This strategy helps prevent early convergence to the uniform distribution for misclassified samples, allowing the network to potentially classify them correctly in subsequent training iterations.

\subsection{Evidential Regression}
 
Thus far, our discussion has exclusively centered around classification problems. However, when it comes to predicting continuous targets, such as in regression problems, the concept of the ``I don't know'' outcome is not explicitly defined because the range of possible outcomes is unbounded. Subjective logic is primarily tailored to handle uncertainty in categorical or discrete scenarios, where belief masses and degrees of uncertainty can be assigned to various hypotheses or propositions. \citet{amini2020} approached the regression problem similarly to \citet{sensoy2018evidential} but adjusted the formulation to handle continuous outcomes by changing both the target distribution and the loss function. 

In the context of regression tasks, we work with a dataset $\mathcal{D}$ = $\{x_i, y_i\}_{i=1}^{N}$ where each of the targets $y$ are assumed to be independently and identically distributed (i.i.d.) and drawn from a Gaussian distribution characterized by an unknown mean $\mu$ and variance $\sigma^2$ \citep{gelman1995bayesian}. The deterministic model shown in Figure~\ref{fig:models}(b)(i) predicts just $\hat{y}$ and is trained using an expected value loss function such as the mean-squared error. Figure~\ref{fig:models}(b)(ii) shows a parametric model for predicting parameters $\mu$ and variance $\sigma^2$. Here, as with the multinomial distribution for categorical problems, if maximum likelihood is used to derive an optimization loss, the likelihood parameters $\mu$ and $\sigma^2$ are treated as fixed and deterministic, hindering the estimation of epistemic uncertainty. \citet{amini2020}'s evidential regression assumes the mean $\mu$ is drawn from a Gaussian distribution, while the variance $\sigma^2$ follows an Inverse-Gamma distribution
\begin{eqnarray}
y_n \sim \mathcal{N}(\mu_n, \sigma_n^2) \\ 
\mu_n \sim \mathcal{N}(\gamma_n, \sigma_n^2 \nu_n^{-1}) \\ 
\sigma_n^2 \sim \Gamma^{-1} (\alpha_n, \beta_n),
\end{eqnarray}
where $\mathbf{m}_n = (\gamma_n, \nu_n, \alpha_n, \beta_n)$, $\gamma_n \in \mathbb{R}$, $\nu_n>0$, $\alpha_n>1$, and $\beta_n>0$. This formulation results in a higher-order distribution referred to as the ``evidential distribution'', denoted as $p(\mu_n, \sigma_n^2|\mathbf{m})$, which can be represented by a Normal-Inverse-Gamma distribution 
\begin{eqnarray}
\label{nig_dist}
p(\mu_n, \sigma_n^2 | \gamma_n, \nu_n, \alpha_n, \beta_n) &= \frac{{\sqrt{\nu_n}}}{{\sqrt{2\pi\sigma_n^2}}} \frac{{\beta^\alpha_n}}{{\Gamma(\alpha_n)}} \left(\frac{1}{{\sigma_n^2}}\right)^{\alpha_n+1}\exp\left(-\frac{{2\beta_n + \nu_n(\mu_n - \gamma_n)^2}}{{2\sigma_n^2}}\right) \\ 
&= \text{St}\left(\gamma_n, \frac{\beta_n(1+\nu_n)}{\nu_n \alpha_n}, 2 \alpha_n \right),
\end{eqnarray}
where St is the Student-t distribution evaluated with location $\gamma_n$, scale $\frac{\beta_n(1+\nu_n)}{\nu_n \alpha_n}$, and $2 \alpha_n$ degrees of freedom. For multi-task models, there are four parameters for each model target, as is shown in Figure~\ref{fig:models}(b)(iii). By learning the parameters $\mathbf{m}$ through training, evidential deep learning models define full distributions over the likelihood parameters $(\mu_n, \sigma_n^2)$, allowing for a comprehensive representation of uncertainty in the model's predictions. The loss used to train a parametric neural network for predicting $\mathbf{m}$, is computed by taking the negative logarithm of Equation~\ref{nig_dist}
\begin{equation}
\mathcal{L}_n^{NLL}(\mathbf{w}) = \frac{1}{2} \log\left(\frac{\pi}{\nu_n}\right) - \alpha_n \log(\Omega_n) + \left(\alpha_n + \frac{1}{2}\right) \log\left((y_n-\gamma_n)^2 \nu_n + \Omega_n\right) + \log\left(\frac{\Gamma(\alpha_n)}{\Gamma(\alpha_n + \frac{1}{2})}\right),
\end{equation}
where $\Omega_n = 2 \beta_n(1 + \nu_n)$. Following the approach employed by \citet{sensoy2018evidential}, \citet{amini2020} included an additional regularizer term to suppress evidence (or raise the uncertainty) in support of incorrect predictions
\begin{equation}
\label{surpress_term}
    \mathcal{L}_n^R(\mathbf{w}) = |y_n - \gamma_n| (2 \nu_n + \alpha_n),
\end{equation}
where the first term represents the model error while the second term is proportional to the total evidence accumulated by the learned posterior. The total loss used during training is finally
\begin{equation}
\label{ev_reg_loss}
    \mathcal{L}_n(\mathbf{w}) = \mathcal{L}_n^{NLL}(\mathbf{w}) + \lambda \mathcal{L}_n^R(\mathbf{w}),
\end{equation}
where the parameter $\lambda$ is selected to best calibrate the model. As we show in the results section (and as others have shown), if $\lambda$ is too small, the model tends to over-fit the data, while overly large values of $\lambda$ lead to uncertainty over-inflation \citep{soleimany2021evidential}.

\subsection{Law of Total Variance and Entropy}

The law of total variance \citep[LoTV;][]{casella2002statistical} states that for two random variables $X$ and $Y$ on the same probability space, the variance of variable $Y$ may be decomposed as 
\begin{equation}
\label{lotv}
\text{Var}(Y) = \mathbb{E}[\text{Var}(Y|X)] + \text{Var}(\mathbb{E}[Y|X]),
\end{equation}
where $\text{Var}(Y)$ represents the total variance of the random variable $Y$, $\mathbb{E}[\text{Var}(Y|X)]$ denotes the expected value of the conditional variance of $Y$ given $X$, and $\text{Var}(\mathbb{E}[Y|X])$ represents the variance of the conditional means of $Y$ given $X$. The first term is often referred to as the ``aleatoric'' uncertainty or ``uncertainty in the data,'' while the latter represents the ``epistemic'' uncertainty or ``uncertainty in the model's predictions.''

For the Dirichlet distribution, these quantities can be computed as 
\begin{eqnarray}
\underbrace{\mathbb{E}[\text{Var}(y_{n,k}|p_{n,k})]}_\text{Aleatoric} &= \mathbb{E}[p_{n,k}(1-p_{n,k})]  \\
\label{lotv-ale}
&= \frac{\alpha_{n,k}}{S} - \left( \frac{\alpha_{n,k}}{S} \right)^2 - \frac{\frac{\alpha_{n,k}}{S}\left(1 - \frac{\alpha_{n,k}}{S} \right)}{S + 1} \\ 
\underbrace{\text{Var}(\mathbb{E}[y_{n,k}|p_{n,k}])}_\text{Epistemic} &= \text{Var}(p_{n,k})  \\ 
\label{lotv-epi}
&= \frac{\frac{\alpha_{n,k}}{S}\left(1 - \frac{\alpha_{n,k}}{S} \right)}{S + 1}.
\end{eqnarray}
The epistemic uncertainty computed with the LoTV and the quantity $u$ from DST do not have the same form. 

For the Normal-Inverse Gamma distribution, we have 
\begin{eqnarray}
\underbrace{\mathbb{E}[\mu_n] = \gamma_n}_\text{Prediction} \\  
\underbrace{\mathbb{E}[\sigma_n^2] = \frac{\beta_n}{\alpha_n - 1}}_\text{Aleatoric} \\ 
\underbrace{\text{Var}[\mu_n] = \frac{\beta_n}{\nu_n (\alpha_n-1)}}_\text{Epistemic}
\end{eqnarray}
Note the relationship between the two uncertainty quantities: the epistemic uncertainty is the aleatoric uncertainty divided by parameter $\nu_n$, which is interpreted as a ``virtual observation count'' that controls how strongly the prior influences the posterior to the evidence acquired from the data. As $\nu_n$ increases, the prior dominates, so the data must be very persuasive to move the posterior away from the prior prediction $\gamma_n$. With lower $\nu_n$, the data readily overrides the prior, allowing the posterior to diverge from $\gamma_n$. So $\nu_n$ modulates the impact of the data versus the prior, rather than literally counting observations.

The LoTV is not limited to stochastic models; it can also be applied to  ensembles of probabilistic models. For instance, in the case of a Gaussian parametric model that predicts ($\mu_n$, $\sigma_n^2$), an ensemble can be created using various approaches discussed below. Conversely, in a categorical problem, the variance can be estimated directly from the predicted probabilities and the true labels, as shown in Equation~S5. Given an input $\mathbf{x}_n$ to the ensemble of trained models, the output consists of a list of predicted means and variances. By using Equation~\ref{lotv}, we can compute the aleatoric and epistemic components.

In ensembles involving categorical models, additional statistical quantities can be computed alongside Equation~\ref{lotv}. One such quantity is entropy, denoted as $H$, which serves as a measure of uncertainty encompassing both the data distribution and the model's predictions. The formulation for entropy, as given by Equation~\ref{entropy}, is expressed as follows:

\begin{eqnarray}
p_k = \frac{1}{E} \sum_{e=1}^{E} p_{e,k} \\
\label{entropy}
H = - \sum_{k=1}^{K} p_k \log(p_k),
\end{eqnarray}

In this equation, $p_j$ represents the ensemble average over $E$ ensemble members. The range of values for entropy spans from 0 to $\log(K)$, where 0 signifies perfect certainty (indicating that only one class has a probability of 1), and $\log(K)$ signifies maximum uncertainty (implying that all classes have equal probabilities). 

\subsection{Evaluation Metrics}

To quantitatively assess the performance and uncertainty of our models trained on categorical (precipitation type) and regression (SL) datasets, we adopt the evaluation methods outlined by \citet{haynes2023creating}. This assessment framework centers around four key objectives described below. First, the accuracy of mean predictions generated by our models is prioritized and relies on metrics such as the Brier Skill Score (BSS) and Root Mean Square Error (RMSE). These metrics, visualized through the attributes diagram, help to gauge how well each model produces an accurate and calibrated mean forecast. 

For classification problems, the Brier score \citep[BS;][]{Brier1950-af} measures the mean squared difference between the observed label ($y_{n,k}$) and the predicted probabilities ($p_{n,k}$) given by
\begin{equation}
\text{BS} = \frac{1}{N K} \sum_{n=1}^{N} \sum_{k=1}^{K} (y_{n,k} - p_{n,k})^2,
\end{equation}
where $N$ is the total number of events and $K$ is the total number of outcomes. A lower Brier score is better with a minimum possible score of 0. We also utilize the Brier Skill Score (BSS), which is computed as $BSS$ = $1 - \frac{BS_{\text{forecast}}}{BS_{\text{climatology}}}$, where $BS_{\text{forecast}}$ is the Brier score of the predicted forecast, while $BS_{\text{climatology}}$ is the Brier score of the climatological forecast, which is the mean squared difference between the observed frequency of the event and the predicted probability of the event based on climatology (the long-term historical frequency of the event). The BSS ranges from negative infinity to 1. A BSS of 1 indicates a perfect forecast (no error), and a negative BSS suggests that the forecast is worse than the climatology. A BSS of 0 means the forecast is as skillful as climatology.

In the regression problem, we employ the root mean square error (RMSE) as the performance metric, defined as
\begin{equation}
\text{RMSE} = \sqrt{\frac{1}{N}\sum_{i=1}^{N}(y_i - \hat{y}_i)^2},
\end{equation}
where $y_i$ represents the ground truth labels and $\hat{y}_i$ denotes the model predictions.

Secondly, we quantify the calibration of uncertainty estimates, ensuring that predicted uncertainties closely align with observed outcomes. This alignment is assessed using the Probability Integral Transform (PIT) plot, which assesses if the model produces unbiased and well-calibrated distributions. The PIT represents the quantile of the predicted distribution at which the observed value occurs. It is calculated by approximating the cumulative distribution function (CDF) of the predicted distribution and evaluating it with the observed value. The PIT histogram displays the distribution of PIT values over multiple data samples, aiming for a uniform histogram in a perfectly calibrated model. The deviation score for the PIT \citep{nipen2011calibrating} can be computed as
\begin{equation}
\text{PITD} = \sqrt{\frac{1}{M} \sum_{i=1}^{N_M} \left(\frac{N_M}{N} - \frac{1}{M}\right)^2}.
\end{equation}
In this formula, $M$ represents the number of bins used for the PIT histogram, $N_M$ is the count of samples in each bin, and $N$ denotes the total number of samples. The PITD value ranges between 0 and 1, with 0 indicating the best possible calibration. However, the score is sensitive to the number of bins used for the evaluation. To control for this sensitivity, we evaluate the calibration performance in comparison to the worst-case scenario, we define the PITD Skill Score as follows
\begin{equation}
\text{PITD Skill Score} = 1 - \frac{\text{PITD}}{\text{PITD}_{\text{worst}}},
\end{equation}
where $\text{PITD}_{\text{worst}}$ represents the worst possible PITD score, which assumes that all the forecasts end up in one of the outermost bins. The PITD Skill Score quantifies how well the probabilistic forecasts align with the observed probabilities, with higher scores (1 being the best) indicating better calibration relative to the worst-case scenario. Both a uniform PITD and a 1-1 relationship in the spread-skill diagram indicate that a model is calibrated according to uncertainty, and not a uniform PITD alone \citep{hamill2001interpretation}.

Thirdly, we investigate discrimination performance, particularly concerning the confidence levels in a models' predictions. The discard fraction plot aids in understanding how the model's ability to differentiate between various outcomes improves as prediction confidence grows, involves sorting the data by uncertainty and progressively removing data points with higher uncertainties to assess any improvement in model performance \citep{durr2020probabilistic, barnes2021controlled}. This test provides insights into the correlation between prediction error and uncertainty, highlighting whether excluding instances with high uncertainty enhances model performance. In operational scenarios, the discard test is particularly valuable for setting model spread thresholds based on acceptable error levels.

Lastly, in the regression problem we calculate the dependency of the RMSE on the predicted spread as an alternative to traditional error metrics, offering a means to normalize model performance against a baseline. This approach, depicted in the skill-score diagram, helps to provide further insights into the relationship between RMSE and forecast spread \citep{delle2013probabilistic}.

\section{Results}

\subsection{Winter Precipitation-Type}

\begin{figure}[t!]
    \centering 
    \includegraphics[width=\columnwidth]{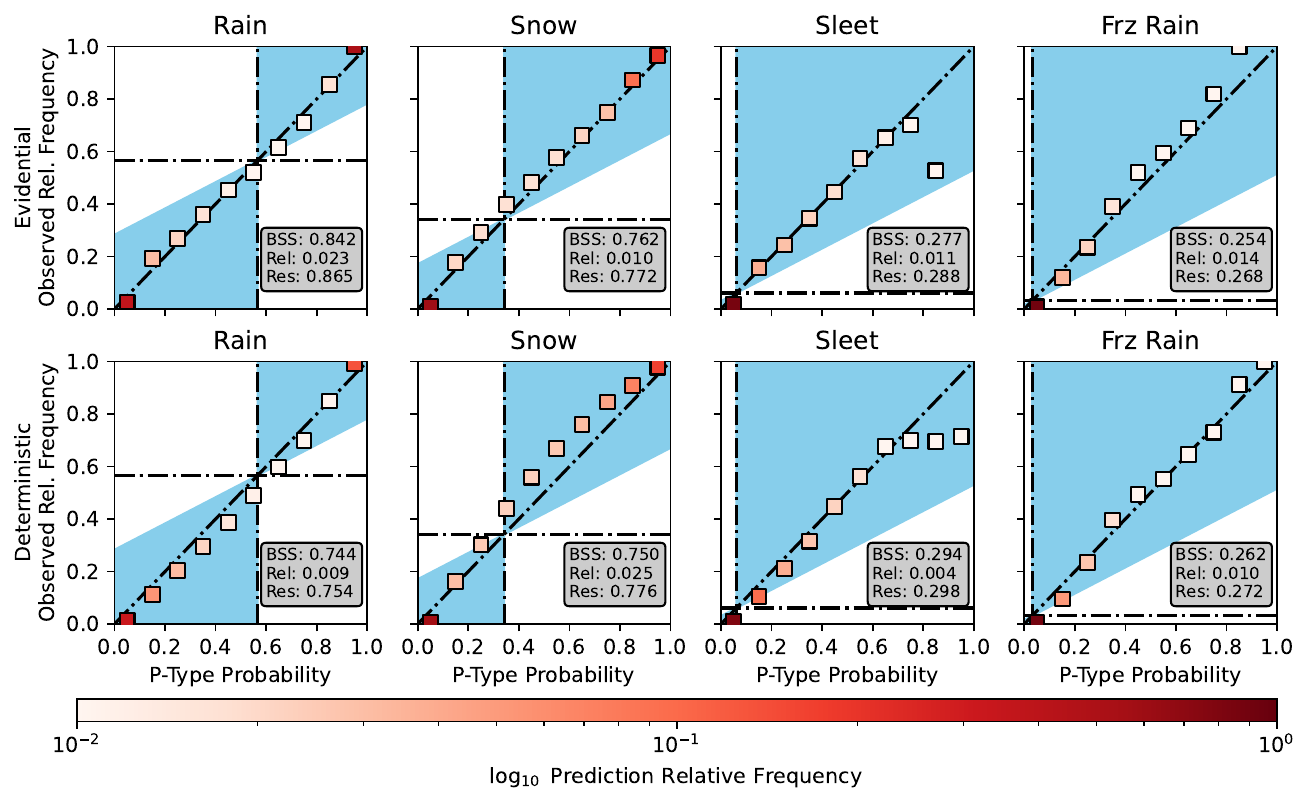}
    \caption{Attributes diagrams for the (a) the evidential model and (b) a deterministic neural network model. The columns show the result for each precipitation-type. In each sub-panel, the diagonal, horizontal, and vertical dashed lines indicate the 1-to-1 line, no-resolution line, and climatology line. The blue-shaded area indicates skill relative to climatology. Red-shaded rectangles illustrate the reliability of each model in predicting each class. The legend in each panel displays the Brier Skill Score, along with the reliability (Rel) and resolution (Res) components of the Brier score scaled by the uncertainty component. Reliability describes the deviation of the predicted probability from the observed relative frequency (lower is better) and Resolution describes the average difference in the predicted probabilities from climatology, related to sharpness where higher values are better.}
    \label{fig:attributes}
\end{figure}

Figure~\ref{fig:attributes} presents a comparison between a deterministic MLP in (a) and an evidential MLP in (b) using reliability curves on attributes diagrams. Both models demonstrate similar reliability for all precipitation-types. The evidential model outperforms the deterministic MLP on rain and snow but performs slightly worse on sleet and freezing rain. Overall the model performs the best on rain, then snow, freezing rain, and finally sleet. These results are summarized by the confusion matrix plotted in Figure~S1 for both deterministic and evidential MLPs, which again show very similar performance between the two modeling approaches.

The attributes diagram shows that both models produce well-calibrated probabilistic predictions for all precipitation types. Sleet exhibits an overconfident deviation at high probabilities in both models. This issue may be attributed to inconsistent labels in the mPING dataset that were not captured by the wet-bulb temperature filter. Some instances of wet snow may be mistakenly labeled as sleet. While not directly addressing data quality issues here, this finding provides an opportunity to explore the effectiveness of computed and predicted uncertainties when dealing with noisy data.

\begin{figure}[t!]
    \centering
    \includegraphics[width=\columnwidth]{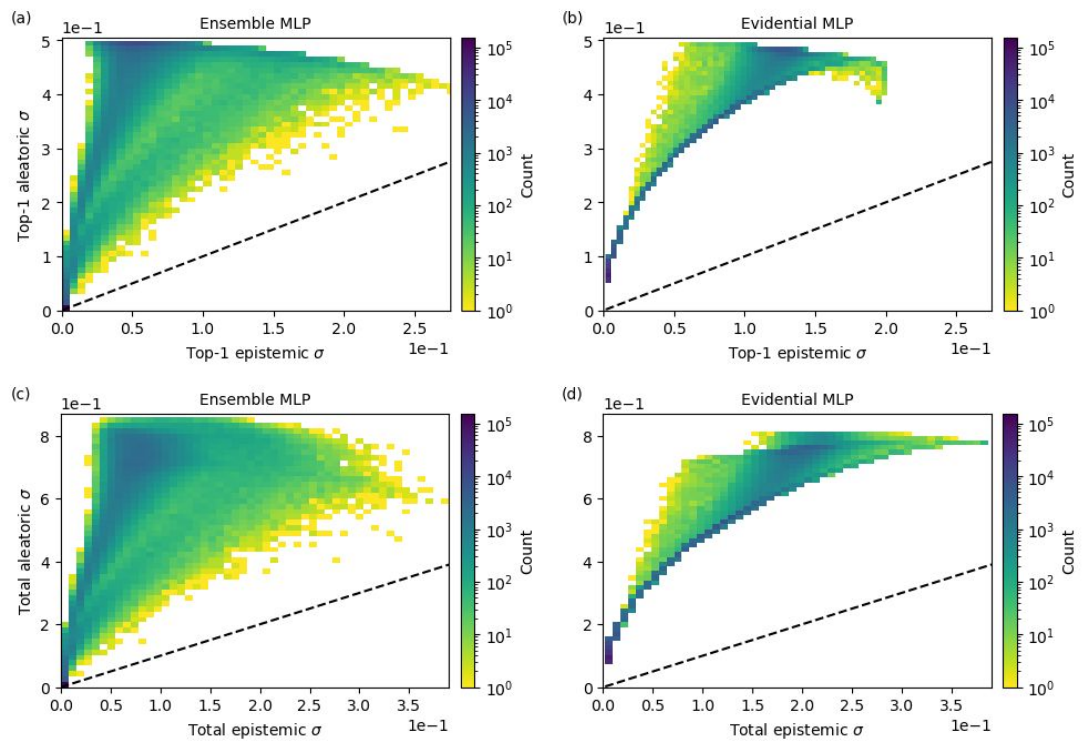}
    \caption{2D histograms illustrate the relationship between the top-1 aleatoric and top-1 epistemic uncertainties for (a) the MC-ensemble deterministic model and (b) the evidential model. In (c) and (d), the total summed quantities for these uncertainties are compared.}
    \label{fig:top1_uq}
\end{figure}

Figures~\ref{fig:top1_uq}(a) and \ref{fig:top1_uq}(b) compare aleatoric and epistemic uncertainties for the true class (referred to as `top-1') in an Monte Carlo ensemble of deterministic MLPs and an evidential MLP model, respectively. Figures~\ref{fig:top1_uq}(c) and \ref{fig:top1_uq}(d) display the summed variances across the four precipitation-types. Throughout the analysis, the predicted aleatoric uncertainty consistently surpasses the predicted epistemic uncertainty (see x/y tick magnitudes), indicating higher variability in the RAP and mPING training datasets compared to the model fit. 

The histograms in Figures~\ref{fig:top1_uq}(b) and (d) are shaped using Equations~\ref{lotv-ale} and \ref{lotv-epi}, respectively. The ensemble MLP's histogram shape resembles that of the evidential model but appears more spread out, suggesting possible stochastic approximation in the ensemble model. On the other hand, the results in Figure~\ref{fig:top1_uq}(b) and (d) are computed using derived expressions. Comparing the ensemble MLP to the evidential MLP, the top-1 versus the summed distributions show greater similarity in the ensemble model, while the evidential model exhibits a more complex relationship, in contrast to the generally monotonic behavior observed in the summed distribution (see below).

\begin{figure}[t!]
    \centering
    \includegraphics[width=\columnwidth]{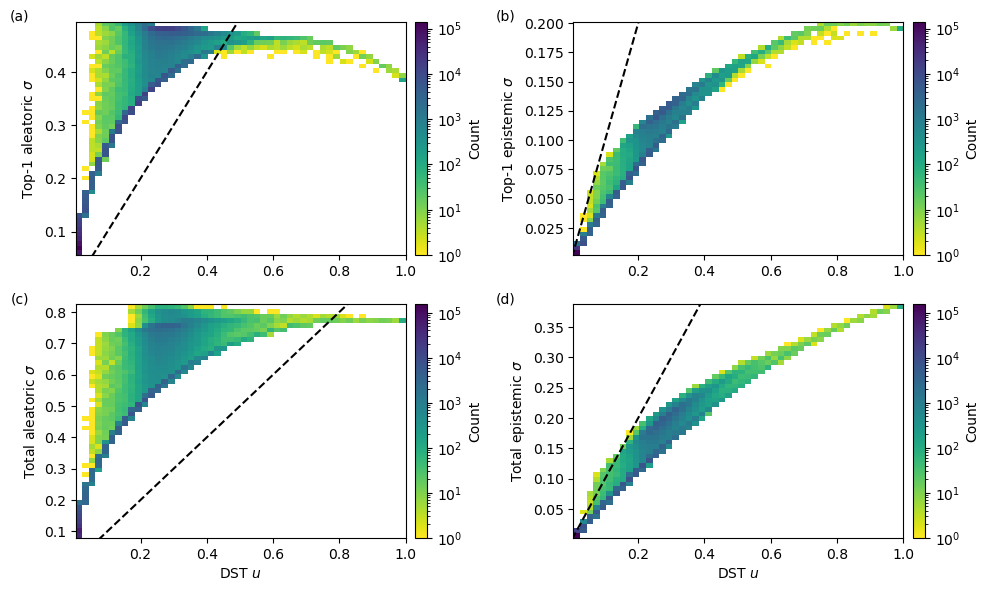}
    \caption{2D histograms compare aleatoric and epistemic uncertainty estimates computed using the LoTV and evidential $u$ from DST. Figures (a) and (b) plot the top-1 predicted aleatoric and epistemic uncertainties, respectively, against $u$. Figures (c) and (d) show the total aleatoric and total epistemic uncertainties, summed over all classes, versus $u$. The dashed line indicates a 1-to-1 relationship in each panel.}
    \label{fig:lotv_vs_ds}
\end{figure}

Figure~\ref{fig:lotv_vs_ds} compares the aleatoric and epistemic contributions to the total uncertainty according to the LoTV, and DST estimates of uncertainty. The LoTV quantifies variance for each precipitation-type, while DST produces a single uncertainty value $u$ across all classes per input $x$. Although both aim to measure model prediction uncertainty, they differ in focus. The LoTV targets model limitations through epistemic variance. In contrast, DST reflects uncertainties in the evidence itself and resulting belief in hypotheses, rather than just model shortcomings. Using both provides comprehensive insights into uncertainty sources.

Figures~\ref{fig:lotv_vs_ds}(a) and (c) show the aleatoric uncertainty exceeds DST $u$ for most events. This result implies that there is strong evidence for at least one of the predicted precipitation types but that in many cases one cannot be unambiguously determined due to the inherent noisiness of the relationship between the RAP profile and the observed precipitation type. As a numerical weather model, RAP has unavoidable errors from numerical discretization approximations and sub-grid parameterizations. Its coarse spatiotemporal resolution also likely misses some atmospheric and terrain variations, further inflating aleatoric uncertainty.

When aleatoric uncertainty dominates, potential mPING label errors may not substantially impact epistemic or DST uncertainties. In this regime, model uncertainties stem more from RAP's variability rather than its own limitations. However, when DST $u$ is relatively high and exceeds aleatoric uncertainty (right of the dashed line), mPING label errors could raise DST uncertainty through inconsistent evidence. Here, truth label uncertainties have greater influence on belief and plausibility functions, increasing DST uncertainty compared to aleatoric.

As $u$ approaches unity, the model becomes more uncertain about any outcome being more likely than any other. Evidence counts even out across classes, observed as flatter summed aleatoric contributions in Figures~\ref{fig:lotv_vs_ds}(a) and (c). With higher DST uncertainty, the model concentrates less high counts in one class, slightly reducing top-1 aleatoric uncertainty.

In Figures~\ref{fig:lotv_vs_ds}(b) and \ref{fig:lotv_vs_ds}(d), the model's epistemic uncertainties are smaller than corresponding DST $u$, as DST allows belief masses to exceed individual class probabilities when considering evidence from multiple sources. In contrast, epistemic uncertainty reflects inherent model limitations and is typically narrower. The monotonic relationship between epistemic and DST $u$ suggests that higher uncertainty about precipitation evidence is a result of the model's limitations. As the DST $u$ value increases, the uncertainty in the model's predictions grows, indicating an increasing lack of confidence in its ability to make precise determinations. The linear relationship between small epistemic and DST $u$ values indicates proportional scaling of lower uncertainties. Larger DST $u$ values arise from its ability to combine evidence from different sources, encompassing more uncertainties than epistemic uncertainty, which solely focuses on model uncertainties. Thus, accounting for multiple uncertainty sources in the model's predictions is crucial.


In the Supplementary Materials, we continue the analysis of model performance and uncertainty, specifically focusing on the comparison between the Brier score and standard deviation (see Figure S2). Both the deterministic MLP and the evidential MLP models generally exhibited aleatoric uncertainty values surpassing the corresponding Brier scores, displaying a sigmoidal correlation between skill and uncertainty. The total uncertainty was comparable between the two approaches, with aleatoric uncertainty exerting a more significant influence than epistemic uncertainty. In the ensemble model, the computed entropy closely mirrored the total uncertainty, primarily driven by aleatoric contributions.

Additionally in the Supplementary Materials, we analyze the relationship between dropout rate and uncertainty calibration for the ensemble of deterministic MLPs and demonstrate the extent to which the dropout rate influences calibration (see Figures S3(a) and S3(b)). Overall, the dropout can be used to fine-tune the calibration, but the role is limited. Additionally, for the evidential model, the loss weight on the KL divergence term (Equation ~\ref{total_loss}) can be used to some extent to adjust calibration for the precipitation-type data set (results not shown). Notably, these hyperparameters' dependence is not very strong, and significant adjustments are generally unnecessary for effective uncertainty calibration. In fact, the uncertainties for the evidential model are predominantly well-calibrated through model optimization on the F1 score alone, without extensive tuning of dropout or loss weighting (see the Supplementary Materials for training and optimization details).

Conversely, calibrating the deterministic model requires the creation of an ensemble of sufficient size to estimate uncertainty distributions, which may grow with the number of classes. As a result, the evidential model offers distinct advantages and is considered the preferred model due to its ability to achieve very similar performance to the ensembles of a deterministic model and well-calibrated uncertainties without the need for complex ensemble techniques or tuning other hyper-parameters \citep{schreckECHO}.

\begin{figure}[t!]
    \centering
    \includegraphics[width=\columnwidth]{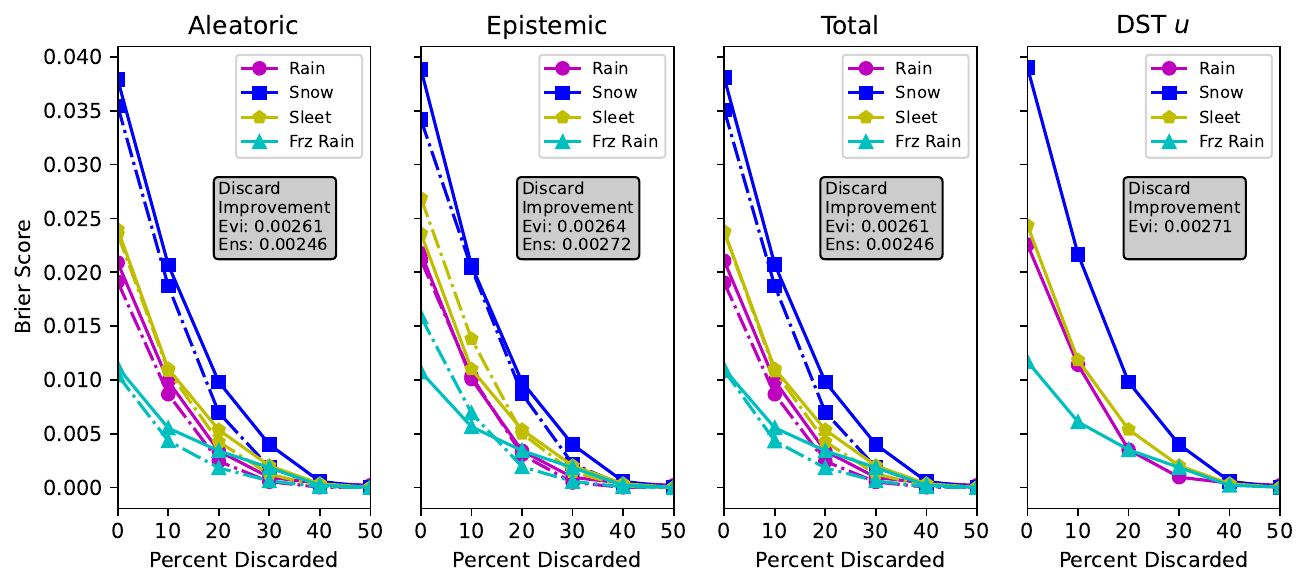}
    \caption{Discard-test diagrams show the fraction of data points removed from the test set versus the BS computed on the remaining subset. Dashed lines illustrate the ensemble model while solid lines show the evidential model. Points above the horizontal dashed (black) line have skill higher relative to climatology. The legend in each panel shows the discard improvement (DI) score, which is a measure of improvement in Brier scores when certain percentiles of the uncertainty metric are discarded. Higher values indicate a better discard improvement.}
    \label{fig:ptype_drop_frac}
\end{figure}

Figure~\ref{fig:ptype_drop_frac} illustrates the discard-test for the ensemble and evidential models across aleatoric, epistemic, total, and entropy/DST uncertainties. In this approach, a threshold is set for the model's predicted uncertainty. During deployment, any predictions above this uncertainty threshold are discarded or considered ``out of confidence.'' Rather than providing potentially misleading or unreliable results in uncertain situations, the model simply abstains from making predictions. This conservative strategy ensures the model does not output predictions when it is uncertain or lacks sufficient evidence.

The BS is used as the performance metric for the discard test. The figure shows that for rain and snow, the BS consistently improves as more uncertain data is withheld from evaluation. Both the ensemble and evidential models exhibit very similar discard test curves for each precipitation type and form of uncertainty. There are slight variations in the Discard Improvement (DI) score with the evidential model having a slightly higher DI than the ensemble for aleatoric but a slightly lower score for epistemic. Each precipitation-type has a different discard slope. Snow has the deepest slope and freezing rain has the shallowest.  
 
\begin{figure}[t!]
    \centering
    \includegraphics[width=\columnwidth]{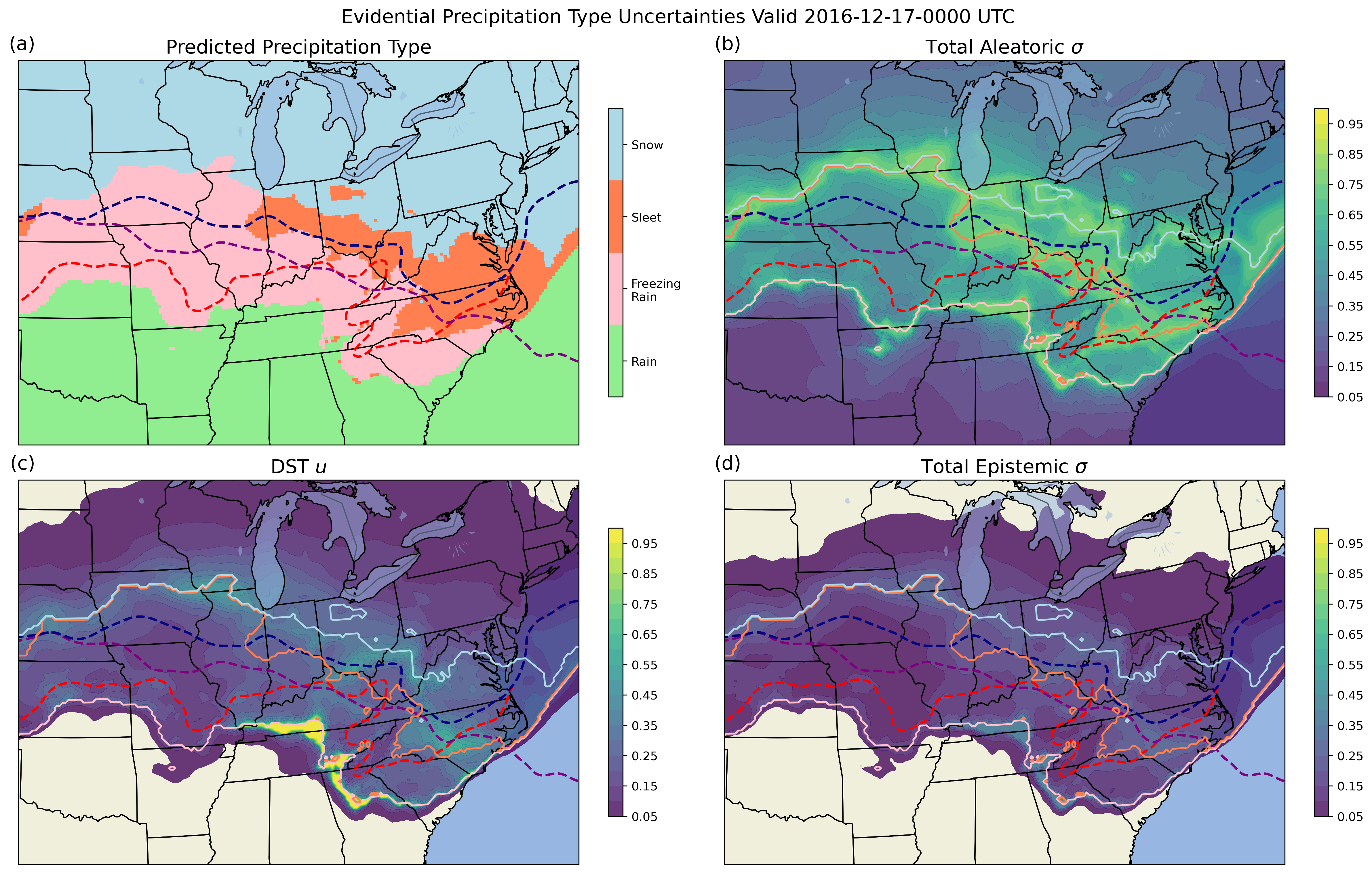}
    \caption{17 December 2016 precipitation-type predictions and their uncertainties from the evidential model are visualized for the central US. Panel (a) displays the most likely precipitation-type prediction, while panels (b-d) showcase the total aleatoric, DST, and epistemic uncertainty, respectively. The red, purple, and navy contours indicate the 0 C isotherm at the surface, 2 km AGL, and the 0-5 km AGL maximum, respectively.}
    \label{fig:conus}
\end{figure}

Figure~\ref{fig:conus} presents a case study of a severe winter weather event that impacted a significant portion of the country on December 17th, 2016. The evaluation of the evidential model was conducted across the entire RAP domain, with the model predicting the probability of each precipitation type conditioned on precipitation occurring. In Figure~\ref{fig:conus}(a), the model's predictions for freezing rain, sleet, snow, and rain are depicted. Freezing rain and sleet are predicted over a vast region, stretching from western Kansas eastward and then southward into South Carolina. Snow is primarily predicted for points north, while rain is predicted for points south, corresponding to the mixed precipitation types.

Figure~\ref{fig:conus}(b) and (c) illustrate the patterns of the predicted aleatoric and epistemic uncertainties, respectively. The highest uncertainties are observed in regions where one precipitation-type transitions into another and along the freezing line. Conversely, predictions of rain and snow are generally accompanied by smaller uncertainties.

Of particular interest is the close correlation between the epistemic uncertainty and the DST $u$ quantity. As mentioned earlier, the DST uncertainty usually has a larger magnitude compared to the epistemic uncertainty. However, Figure~\ref{fig:conus}(d) clearly shows  $u$ spiking to 1 in a relatively small region over northern Georgia stretching up to southern Kentucky, indicating that the model is completely uncertain about those predictions. This situation also corresponds to the highest value of epistemic uncertainty on the grid, showing that both uncertainties align during moments of high uncertainty when the model is unsure about the precipitation-type. Both DST $u$ and total epistemic uncertainty are highest along the rain to freezing rain transition line and lower when transitioning to sleet and snow. This higher epistemic uncertainty may be due both to near-surface temperature uncertainty as well as uncertainty from mPING reporters being able to discriminate between freezing rain and rain consistently.

\begin{figure}[t!]
    \centering
    \includegraphics[width=\columnwidth]{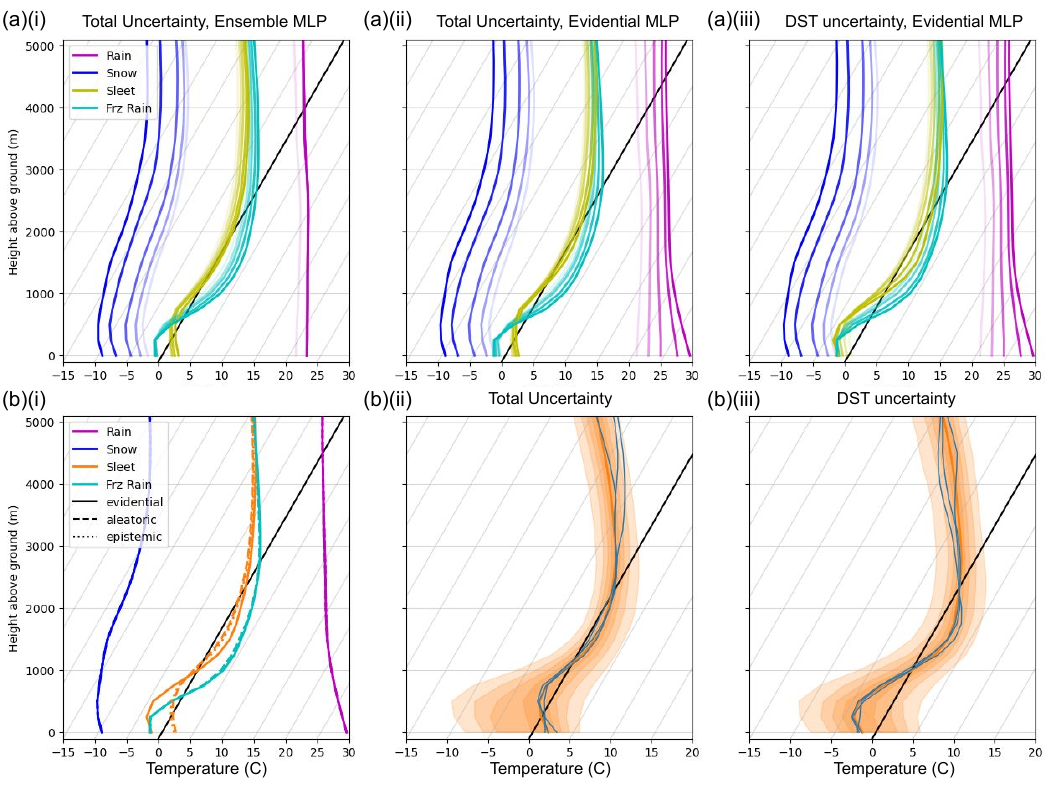}
    \caption{(a) Composite soundings are presented lightest-to-darkest using the median of the 10th, 20th, 50th, and 90th percentile least uncertain predictions generated by the evidential model. The total uncertainty is utilized for the MC-ensemble MLP (i) and the evidential MLP (ii), while the DST uncertainty is employed for the evidential MLP in (iii). (b)(i) showcases the breakdown of the top 10\% least uncertain composites (depicting the median) based on class and uncertainty. Panels (b)(ii-iii) illustrate the densities of the 90th percentile least uncertain predictions, using DST $u$ and LoTV total uncertainty, respectively.}
    \label{fig:soundings}
\end{figure}

Lastly, Figure \ref{fig:soundings} shows composite soundings for various models and uncertainty percentiles. The panels in Figure \ref{fig:soundings}(a)(i-iii) demonstrate that lower uncertainty predictions correspond with more prototypical soundings for the corresponding prediction type.
DST uncertainty shows the best performance since the ground level temperature of the composite sleet soundings remains below the freezing line. 
The other figures, for the MC ensemble and total uncertainty, have an above-freezing composite at the ground level.

The discrepancies between the different types of uncertainty for the evidential neural network are exhibited by Figure \ref{fig:soundings}(b)(i). In this figure, only sleet has varying composite soundings between the different types of uncertainty (for the 90th percentile of least uncertainty predictions). 
This occurs because even if a prediction is in the 90th percentile of epistemic or aleatoric uncertainty, the other uncertainty type could still be high. 
As noted earlier, the aleatoric uncertainty typically dominates the epistemic uncertainty, and thus is why the median sleet sounding of the total uncertainty in Figure \ref{fig:soundings}(a)(iii) is close to the median aleatoric/epistemic sounding in Figure \ref{fig:soundings}(b)(i).
For the density plots in Figures~\ref{fig:soundings}(b)(ii-iii), which show sleet, while the range of the soundings is about equal, the distribution using DST uncertainty is skewed more negative than using total uncertainty. Overall the DST uncertainty seems to identify more physically reasonable soundings for sleet. 

\subsection{Surface layer flux problem}

\begin{figure}[t!]
    \centering
    \includegraphics[width=\columnwidth]{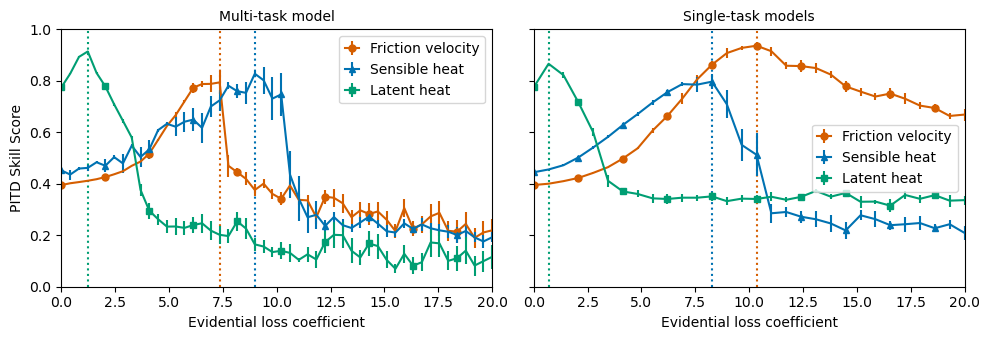}
    \caption{The PITD score as a function of the evidential loss coefficient ($\lambda$, Equation~\ref{surpress_term}) for (a) one multi-task model and (b) three single-task models.}
    \label{fig:regression_calibration}
\end{figure}

In this section, we present a similar evaluation of the evidential regression model trained on the SL data. The key point of comparison, detailed below, is that while the evidential regression model provides a solution for modeling uncertainties like the categorical approach, its calibration is much more sensitive to the weight $\lambda$ multiplying the KL term in Equation~\ref{ev_reg_loss} during training. Furthermore, we observe the regression model infrequently producing unrealistic uncertainties, even when calibrated, potentially posing operational challenges. In addition to the evidential regression results, we also present results from ensembles created using cross-validation, deep ensembling, and Monte Carlo dropout for comparison.

We initially address the issue of model calibration by examining Figure~\ref{fig:regression_calibration}(a), which displays the PITD score as a function of the KL weight ($\lambda$) for a three-task model that predicts fraction velocity, latent heat, and sensible heat. The dataset's characteristics prevent the model from achieving the maximum PITD score at the same $\lambda$ value for all three tasks, and none of the tasks approached a perfect skill score of one. To further explore this relationship, we analyze Figure~\ref{fig:regression_calibration}(b), which depicts the same dependency computed using three single-task models. Effective utilization of the evidential regression model on the SL dataset requires three single-task models to determine the optimal $\lambda$ values (illustrated in the figure), and furthermore, the values differ relative to the three-task model shown in Figure~\ref{fig:regression_calibration}(a). Recall that a flat PIT histogram (PITD equal to zero) does not necessarily guarantee calibration \citep{chapman2022probabilistic, haynes2023creating, hamill2001interpretation}.

\begin{figure}[t!]
    \centering
    \includegraphics[width=\columnwidth]{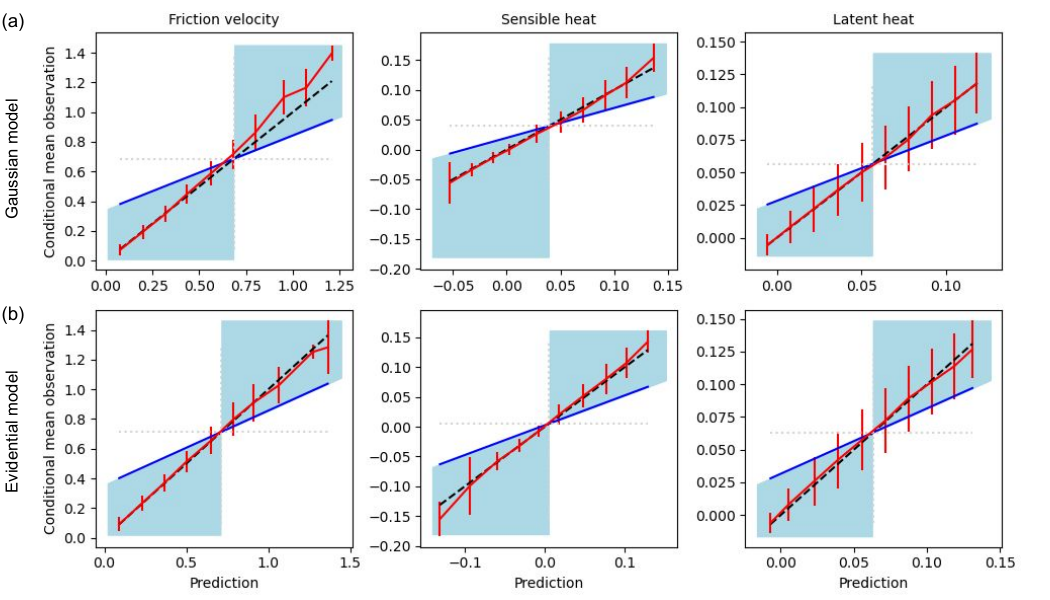}
    \caption{Attribution curves for the (a) data ensemble using the parametric Gaussian model and (b) the evidential model. The red line shows the reliability curve. The columns show the result for each model task. In each subpanel, the diagonal, horizontal, and vertical dashed lines indicate the 1-to-1 line, no-resolution line, and climatology line. The blue-shaded area indicates skill relative to climatology.}
    \label{fig:sl_attribute}
\end{figure}

With the calibration weights identified, the three models were trained and compared against an ensemble of Gaussian models created using cross-validation splits. Figure~\ref{fig:sl_attribute} displays the attributes diagram for the evidential model and the Gaussian model (the results for other ensembles are shown in Figure S5). Overall, both models show similar performance, although there are some discrepancies, such as the lower performance of the Gaussian model, which still remains within the skill region depicted in the figure. Thus, similar to the categorical model, the evidential variation on the standard MLP for the regression problem appears to be an effective substitute for the principal prediction tasks at hand. 

\begin{figure}[t!]
    \centering
    \includegraphics[width=\columnwidth]{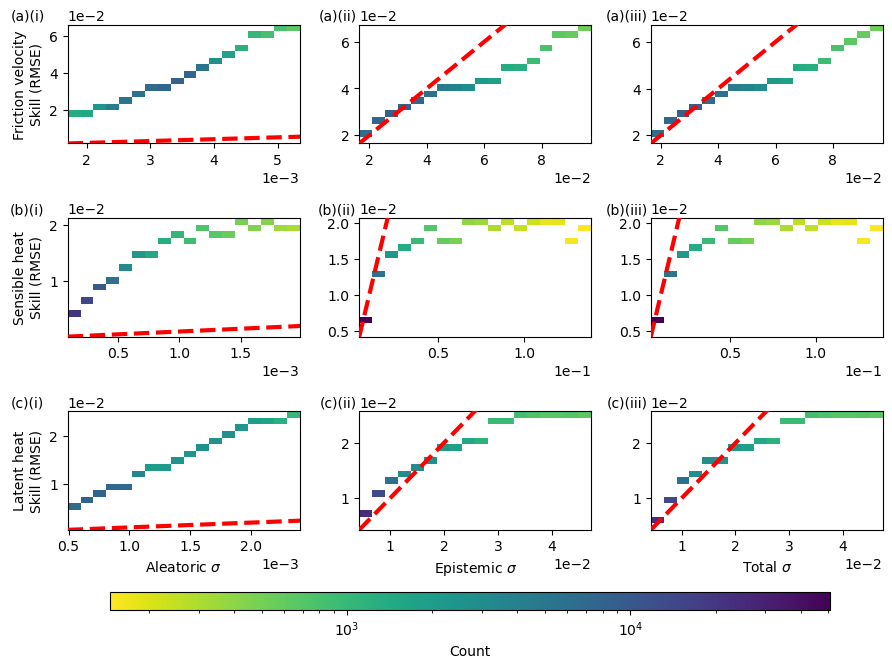}
    \caption{The spread-skill relationship is depicted using a 2D histogram, illustrating the relationship between the standard deviation ($\sigma$) and RMSE for the three single-task evidential models. Each row in (a-c) represents a specific model task, while columns (i-iii) display the results for aleatoric, epistemic, and total uncertainties. Each subpanel features a 1-to-1 line (dashed) and all panels share the color bar indicating the count of each 2D bin.}
    \label{fig:sl_spread}
\end{figure}

How effective is the calibration? Figure~\ref{fig:sl_attribute} shows 2D histograms quantifying the relationship between the computed RMSE for the three model tasks and aleatoric, epistemic, and total uncertainty. Unlike the precipitation-type problem, the dominant uncertainty is epistemic uncertainty which contributes the most to the total uncertainty. The 1-1 relationship is observed for epistemic and total uncertainties when both the RMSE and relative uncertainty are generally small, and this relationship is observed for more than half of the testing data in each case. However, as the quantities increase, the computed RMSE flattens out for sensible and latent heat, while for friction velocity, the relationship appears to be linear with an initial flattening of the RMSE for relatively higher values of uncertainty, which then continues to grow linearly. Note also that none of the models were calibrated according to aleatoric uncertainty; in fact, all of them were under-dispersive. Overall, the best PITD skill score and the best fit between the 1-1 line were achieved for friction velocity.

\begin{figure}[t!]
    \centering
    \includegraphics[width=\columnwidth]{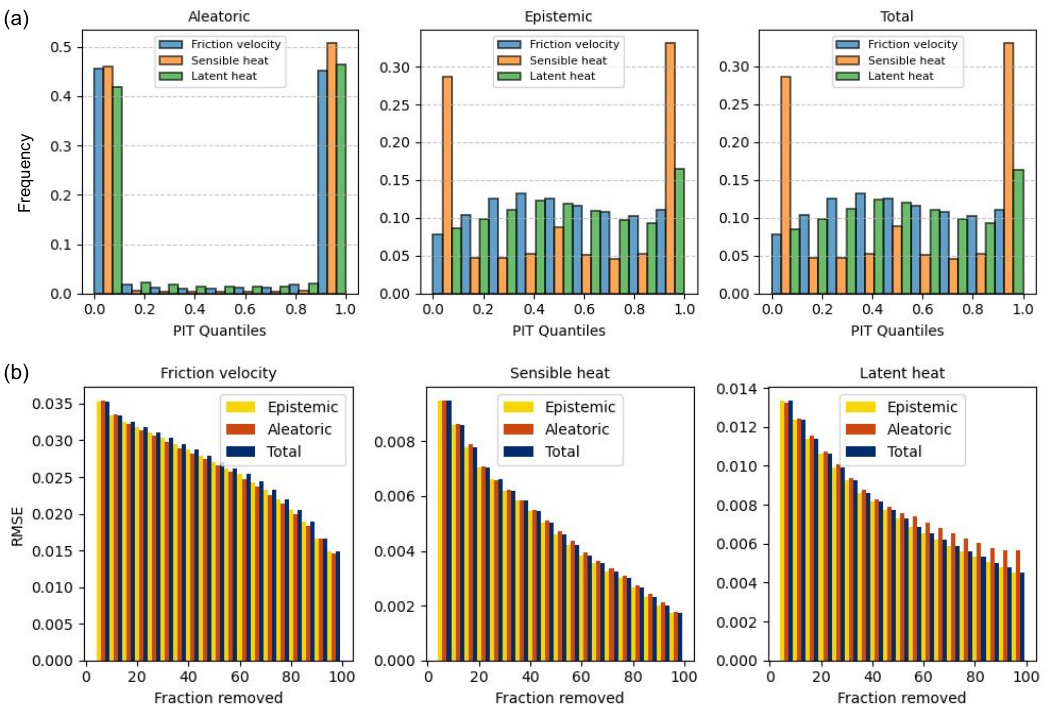}
    \caption{(a) PIT histograms were generated for the three tasks using (i) aleatoric, (ii) epistemic, and (iii) total uncertainty. (b) The discard-test diagrams illustrate the relationship between the fraction of data points removed from the test set and the remaining subset's RMSE for the three single-task models shown in (i-iii).}
    \label{fig:sl_uq}
\end{figure}

These results are further corroborated by the PIT histogram and the discard-test, which are shown in Figures~\ref{fig:sl_uq}(a-b), respectively. The PIT histogram also shows that friction velocity came the closest to calibration, having the flattest histogram of the three models, followed closely by latent heat. Both these quantities show a slight hump towards the middle of the histogram indicating slight under-confidence by the model. Latent heat also has a small hump on the right side of the histogram, further indicating some over-confidence on a subset of the data. The sensible heat model on the other hand clearly shows two humps on the left and right sides of the histogram, showing that the model is overconfident. All of these observations are consistent with Figure~\ref{fig:sl_spread}. All models were subject to extensive hyper-parameter optimization, which suggests that there are limitations to the degree of calibration possible with the evidential regression approach. 

The discard test shows that for all three models, the predicted aleatoric, epistemic, and total uncertainties are linked to the performance of the model through the RMSE, where the more certain data points have lower RMSE values in each case. This is still the case even though the models possessed different degrees of calibration. Therefore, the uncertainty value can be thresholded such that when the model is in operation, it can be used conservatively when model predictions are too uncertain (e.g. the model will not return a prediction if the predicted uncertainty is larger than the threshold). 

While the above results demonstrate the strong performance of the evidential model, we must emphasize a crucial consideration regarding the actual values of the predicted uncertainties. It is essential to be cautious as they may not consistently align with the range of the (in-distribution) task values \citep{ovadia2019can}. For instance, Figure~S8 illustrates histograms for the best-calibrated friction velocity model. Notice the small hump at the largest values for the predicted epistemic uncertainty, which appears inconsistent with the physical nature of the quantity of interest (and should be relatively straightforward to model). This discrepancy raises a concern about the reliability of uncertainty estimates at extreme values.

Additionally, we conducted tests using a range of other model architectures found through optimization, which demonstrated similar performance to the ones presented here. However, these models produced different ranges for the histogram values at the extreme ends, despite having comparable (but not perfect) calibration. This finding highlights the need for careful interpretation and validation of the uncertainty estimates, as different model architectures can lead to varying distributions of predicted uncertainties, even if the overall calibration is consistent. Therefore, while the evidential model shows promise in terms of performance, it is crucial to critically evaluate and validate the predicted uncertainties to ensure their alignment with the true nature of the underlying data.

\begin{figure}[t!]
    \centering
    \includegraphics[width=\columnwidth]{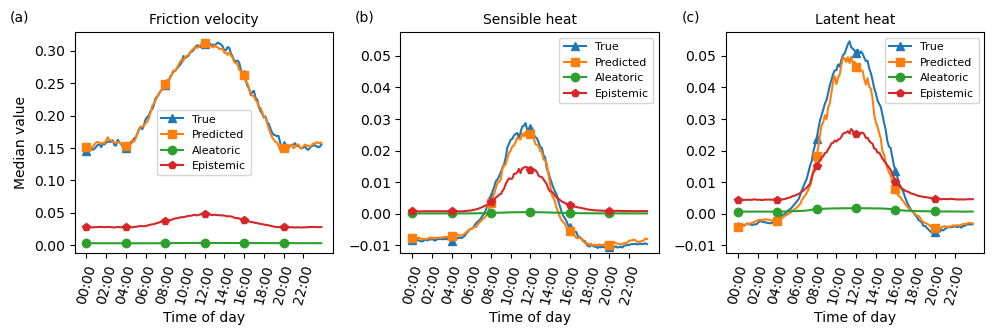}
    \caption{Figure (a-c) illustrates the median values of the friction velocity, sensible heat, and latent heat in 10-minute increments. The curves displayed in each panel represent the truth, predicted, aleatoric, and epistemic quantities. All results are derived from the test data set.}
    \label{fig:daily_SL}
\end{figure}

In our investigation of the SL dataset, we sought to determine whether the predicted uncertainties align with computed errors including the underlying physical processes governing the three tasks. Figure~\ref{fig:daily_SL} presents the computed median predictions and uncertainties as a function of the time of day, with the median helping to mitigate the impact of the outliers on the mean calculation.

The results show a distinct trend in epistemic uncertainty throughout the day. During nighttime, the epistemic uncertainty is relatively low, indicating a higher level of confidence in estimating the quantities. However, as the sun rises and solar radiation intensifies, the epistemic uncertainty rapidly increases, peaking around midday. This rise in uncertainty is attributed to the dynamic and complex atmospheric conditions associated with strong solar heating and convectively-driven atmospheric turbulence during this time. As the day progresses toward night and the sun begins to set, the epistemic uncertainty gradually decreases following the stabilization of the surface layer.

These patterns align with our expectations, as the latent heat, which is influenced by phase changes and processes like evaporation and condensation, exhibits higher uncertainty compared to sensible heat, which primarily relates to temperature changes and has a simpler measurement process. However, even a relatively small change to the value of $\lambda$ can flip this relationship around (Figure~S9). The findings emphasize the importance of selecting an appropriate evidential loss weight to enhance the reliability and usefulness of uncertainty estimates, particularly for latent heat and sensible heat tasks. Proper calibration is crucial to ensuring the model's uncertainty predictions during the daytime remain reliable and physically sensible.

\section{Discussion}

Proper calibration of uncertainties is crucial for regression models employed in applications like weather forecasting \citep{raftery2005using}. However, continuous regression calibration presents distinct challenges compared to classification. A key difficulty is defining calibration to capture whether uncertainties reliably indicate expected errors, as highlighted by \citet{kuleshov2018accurate}. Existing methods exhibit flaws in distinguishing useful and uninformative uncertainties \citep{kuleshov2018accurate}. Furthermore, balancing calibration and the dispersion of predicted probabilities is non-trivial, as techniques like isotonic regression can enforce calibration while increasing dispersion \citep{levi2022evaluating}. Other challenges include modeling complex error distributions \citep{harakeh2020bayesod}, incorporating distinct uncertainties \citep{depeweg2018decomposition}, and ensuring generalization \citep{amershi2019software}. Addressing these issues is imperative for safety-critical systems relying on regression uncertainties.

For categorical problems, DST quantification of uncertainty offers additional benefits. It allows representation and reasoning about uncertain or conflicting evidence, useful in real-world datasets with ambiguous or uncertain ground truth labels. It also enables a principled combination of evidence from multiple sources, beneficial when dealing with different data types or integrating information from diverse sensors or models. The ability of a neural network to parametrically represent partial beliefs is valuable in situations with limited information or incomplete evidence, allowing a more nuanced representation of uncertainty.

The study also demonstrates the potential of evidential deep learning for predictive modeling and uncertainty quantification in meteorological applications. The evidential approach provides calibrated estimates of uncertainty alongside predictions without requiring model ensembles.

For the winter precipitation classification task, the evidential neural network achieves comparable accuracy to standard classifiers while quantifying aleatoric and epistemic uncertainties. The aleatoric uncertainty, driven by inherent randomness in the RAP model which was used as model inputs, generally exceeds the epistemic uncertainty stemming from model generalization errors. This highlights the benefit of decomposing uncertainties to pinpoint dominant sources. In regions with relatively high DST uncertainty, it closely aligns with peaks in epistemic uncertainty, demonstrating how different uncertainty metrics capture related aspects of model confidence.

The discard test validates the use of uncertainty to filter unreliable predictions and improve operational reliability. Setting uncertainty thresholds allows prudent model usage by having it abstain when unsure, preventing potentially misleading high-confidence predictions in uncertain scenarios.

For the surface layer energy flux regression problem, proper calibration is essential for useful uncertainties. While the evidential model's overall performance is just as good as what a deterministic MLP can produce, occasional unrealistic uncertainty values were predicted, emphasizing the need for cautious interpretation due to calibration deficiencies. Architectural choices significantly impact uncertainty characteristics as does the dropout rate. The time-of-day analysis reveals sensible patterns in epistemic uncertainty variation, offering reassurance about reliability. Uncertainty spikes during the daytime when modeling intricacies of turbulent transport and radiation are most challenging. This physics-based behavior appropriately reflects fundamental processes governing the quantities of interest.

A key limitation is the extensive hyperparameter tuning required for uncertainty calibration with the evidential regression model. In contrast, the categorical evidential model provided well-calibrated uncertainties without excessive tuning, highlighting the greater calibration challenges for regression tasks, and motivating continued research. One strategy that we found useful was to first find performant architectures for the relevant prediction tasks, and then tune the loss weight while monitoring the PITD score (this is how Figure~\ref{fig:regression_calibration} was generated) to calibrate as closely as possible each model. 

In general, calibrating ML models to account for epistemic uncertainty requires some kind of calibration/validation dataset or prior assumption derived from similar modeling similar data and cannot be determined alone from the training data. While this is more obvious for post hoc calibration methods such as isotonic regression or conformal prediction, it is also true for evidential methods through the tuning of the KL divergence term in the loss. As a consequence, if the model is applied to a similar dataset with a distribution shift, such as training on analysis data and applying the same evidential model to a forecast, then the evidential uncertainty estimates will be inherently underdispersive. More research is needed to identify ways to adjust uncertainty estimates based on the nature of the domain shift. 

Overall, the study establishes evidential learning as an effective framework combining probabilistic modeling and deep learning strengths. It efficiently provides a means for calibrating predicted uncertainties, addressing standard network limitations, and enhancing interpretability through uncertainty analysis. With further advances, evidential neural networks hold significant potential across meteorological domains.

\section{Conclusions}
In conclusion, this study demonstrates the potential of evidential deep learning as an effective technique for predictive modeling and uncertainty quantification in weather and climate applications, for both classification and regression. The approach synergistically integrates the capabilities of probabilistic modeling with the representational power of deep neural networks. This enables the models to produce well-calibrated estimates of uncertainty alongside predictions, overcoming the limitations of standard deep-learning approaches. The ability to quantify different sources of uncertainty provides valuable insights into model reliability and limitations of the training data.

The decomposition of aleatoric and epistemic uncertainties facilitates detailed analysis to identify challenging prediction cases and opportunities for model improvements. With calibrated uncertainty estimates, evidential neural networks have the potential to enhance understanding of forecast reliability and inform critical decision-making across various meteorological domains, from real-time severe event prediction to long-term climate projections. Given the representational flexibility, computational efficiency, and uncertainty quantification capabilities, evidential deep learning shows promise for tackling a diverse array of prediction and uncertainty estimation problems in the atmospheric and climate sciences.

\section{Software development}

The Machine Integration and Learning for Earth Systems (MILES) group Generalized Uncertainty for Earth System Science (GUESS) package (MILES GUESS) provides tools for estimating and analyzing different sources of uncertainty in earth system science applications. Users working in weather and climate can leverage MILES GUESS to train neural network models that quantify multiple uncertainty types like aleatoric, epistemic, and DST uncertainty.

The package contains layers and losses for evidential deep learning, allowing users to build neural networks that output distributions over targets rather than just point estimates. This enables estimating both aleatoric uncertainty from noise in the data and epistemic uncertainty from model limitations. The code also supports Monte Carlo dropout ensembling for epistemic uncertainty quantification.

Once models are trained with MILES GUESS, the package provides a range of analysis and visualization tools tailored for earth system applications. These include PIT calibration analysis, spread-error diagrams, coverage curves, and more. The code is primarily written in Python and is designed to integrate seamlessly with common earth science workflows based on TensorFlow/Keras and PyTorch. MILES GUESS is accessible at \url{https://github.com/ai2es/miles-guess} where example Jupyter Notebooks demonstrate how to use the package.

\appendix 
\appendixtitle{Supplementary Materials for ``Evidential Deep Learning: Enhancing Predictive Uncertainty Estimation for Earth System Science Applications''}

\setcounter{figure}{0}
\makeatletter 
\renewcommand{\thefigure}{S\@arabic\c@figure}
\setcounter{equation}{0}
\renewcommand{\theequation}{S\@arabic\c@equation}
\setcounter{table}{0}
\renewcommand{\thetable}{S\@Roman\c@table}
\setcounter{section}{0}
\renewcommand{\thesection}{S\@Roman\c@section}

\section{Derivation of the Law of Total Variance}

Let X and Y be random variables defined on the same probability space. The variance of Y can be decomposed into expected values as follows. First, the variance of random variable X is 
\begin{align} 
\label{var_def}
\text{Var}[Y] &= \mathbb{E}[Y^2] - \mathbb{E}[Y]^2 
\end{align}
and the law of total expectation for the random variable $X$ is defined as below:
\begin{align}
\mathbb{E}[Y] &= \mathbb{E} \left[\mathbb{E}[Y|X] \right] 
\end{align}
Next, we apply the law of total expectation and the definition of variance to the first term on the right-hand side of the definition of variance: 
\begin{align*}
\mathbb{E}[Y^2] &= \mathbb{E}[\mathbb{E}[Y^2|X]] \\
&= \mathbb{E}[Var[Y|X] + \mathbb{E}[Y|X]^2]
\end{align*}
Continuing, we apply the law of total expectation to the second term on the right-hand side:
\begin{align}
\mathbb{E}[Y]^2 &= \mathbb{E}[\mathbb{E}[Y|X]]^2
\end{align}
Combining the terms together, we can then rearrange the expectations since the sum of expectations is the expectation of the sums.
\begin{align*}
\mathbb{E}[Y] &=  \mathbb{E}[Var[Y|X]] + \mathbb{E}[Y|X]^2] - \mathbb{E}[\mathbb{E}[Y|X]]^2 \\
&= \mathbb{E}[Var[Y|X]] + (\mathbb{E}[\mathbb{E}[Y|X]^2] - \mathbb{\mathbb{E}[Y|X]}^2)
\end{align*}
Applying the definition of variance to the second part of the term, we get the final formula for the law of total variance:
\begin{align}
\mathbb{E}[Y] &= \mathbb{E}[Var[Y|X]] + Var[\mathbb{E}[Y|X]]
\end{align}

\section{Decomposition for ensembles}
For an ML model predicting a categorical distribution, the expected value for the probability of class $Y_j$ is $p_j$ and the variance is $p_j(1-p_j)$. For an ensemble of K models each predicting a probability distribution across J classes, the total variance of the predictions can be decomposed as:
\begin{align}
Var(Y_j) &= \frac{1}{K} \sum_{k=1}^{K} p_{j,k} (1 - p_{j,k}) + \frac{1}{K}\sum_{k=1}^{K} (p_{j,k} - \overline{p_{j}})^2
\end{align}

\section{Decomposition for evidential classifiers}
For an evidential classifier model using the Dirichlet distribution as its output distribution,
\begin{align*}
\mathbb{E}[Y_j|\textbf{p}] &= p_j\\
Var[Y_j|\textbf{p}] &= p_j(1-p_j)\\
E[p_j] &= \frac{\alpha_j}{S} \\
Var[p_j] &= \frac{\frac{\alpha_j}{S_\alpha}\left(1 - \frac{\alpha_j}{S_\alpha} \right)}{S_\alpha + 1}\\
\mathbb{E}[Var[Y_j|\textbf{p}]] &= \mathbb{E}[p_j(1-p_j)]\\
 &= \mathbb{E}[p_j] - \mathbb{E}[p_j^2] \\ 
 &= \mathbb{E}[p_j] - \mathbb{E}[p_j]^2 - Var[p_j] \\
 &= \frac{\alpha_j}{S} - \left(\frac{\alpha_j}{S}\right)^2 - \frac{\frac{\alpha_j}{S_\alpha}\left(1 - \frac{\alpha_j}{S_\alpha} \right)}{S_\alpha + 1}\\
 Var[\mathbb{E}[Y_j|\textbf{p}]] &= Var[p_j] \\
 &= \frac{\frac{\alpha_j}{S_\alpha}\left(1 - \frac{\alpha_j}{S_\alpha} \right)}{S_\alpha + 1}
\end{align*}

\section{Model Training and Optimization}

The training process for each MLP model (both evidential and deterministic) begins with architecture and trainable weight initialization. The training data is then randomly shuffled. A fixed number of data points called a batch, is selected from the shuffled training data, and the model processes the batch to generate predictions for each point. The mean-absolute error (MAE) was used as the training loss for deterministic regression models, while the relevant evidential loss (categorical or regression) was used for an evidential model. Using the training loss, the model's weights are updated through gradient descent with back-propagation \cite{rumelhart1986}, using a pre-defined learning rate to iteratively reduce the error. This training process, referred to as one epoch, covers the entire training dataset. After each epoch, the training data is again shuffled randomly, and the process is repeated for a specified number of epochs.

To address the class imbalance in the p-type dataset, we utilized two techniques: upsampling and class weights. Upsampling involves artificially increasing the number of samples from under-represented classes, like sleet and freezing rain, by duplicating existing samples. This helps mitigate imbalance and improve model performance on minority classes. Class weights assign higher importance to minority classes during training to counteract the imbalance. Typically these approaches are used individually, but we took a combined approach enabled by hyperparameter optimization (discussed in detail below). We treated class balancing and weighting as tunable parameters, letting the optimizer determine when to balance classes and what weight values to use. This joint optimization of the two strategies produced improved models compared to using either technique alone, for both deterministic and evidential architectures.

The hyper-parameter optimization of all models was carried out using The Earth Computing Hyperparameter Optimization package \cite{schreckECHO} which is built upon Optuna \cite{optuna_2019} and designed to leverage the high-performance computing clusters available at NCAR. The optimization process begins by creating a ``study'' and conducting the first ``trial,'' wherein hyperparameters are chosen from specified ranges, the model is trained, and its performance is evaluated in box simulations. Each trial is independent of others, and trials are executed until the optimization converges or a predetermined number of trials are completed. Upon completion, the study records the outcomes and associated metadata, enabling the estimation of hyperparameter importance for model performance.

For sampling hyperparameters, we employ the Tree-structured Parzen Estimator (TPE) \cite{bergstra2011}. The TPE method fits a Gaussian Mixture model (GMM) $l(x)$ to the best Mean Absolute Error (MAE) values associated with completed trials. A second GMM is also fitted to the remaining parameter values, and the next parameter value is chosen based on maximizing $l(x)$ / $g(x)$. Initially, we use random sampling instead of the TPE sampler to provide observations that the GMM models can leverage for better-informed parameter selections. This step aids in informing the TPE sampler effectively. For further insights into hyperparameter optimization, refer to the Optuna documentation.

All training and hyperparameter optimization configurations (including the best configurations used for the models in this manuscript),  may be found on our GitHub site for the project: \url{https://github.com/ai2es/miles-guess}. For the p-type models, we found that the macro-F1 metric enabled the optimizer to find the ``best'' model for sleet and freezing rain performance while sacrificing some performance in predicting rain and snow. Other metrics including expected and mean calibration error estimates did not perform as well. 

\section{Additional Results}

\begin{figure}[t!]
    \centering
    \includegraphics[width=5 in]{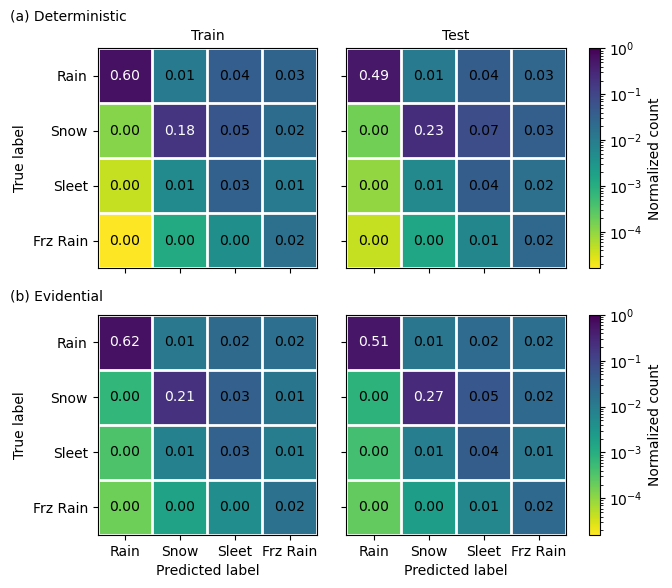}
    \caption{The confusion matrix illustrates the performance of the holdout test split for (a) the deterministic model and (b) the evidential model.}
    \label{fig:s1}
\end{figure}

Figure~\ref{fig:s1} presents a comparison of the performance results between the deterministic and evidential models using a confusion matrix. The confusion matrix is shown as unnormalized to show the model's predictive capabilities and highlights potential classification errors.

\begin{figure}[t!]
    \centering
    \includegraphics[width=\columnwidth]{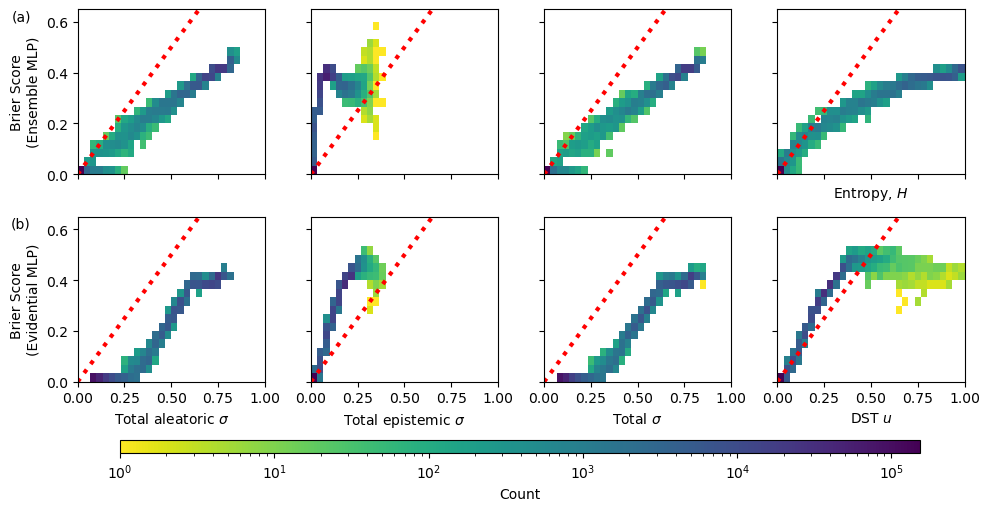}
    \caption{2D histograms illustrate the relationship of Brier score versus standard deviation computed using the total (summed) aleatoric, epistemic, and total uncertainties, as well as DST/entropy estimates. In (a) the MC-ensemble deterministic results are shown, while (b) shows the evidential model. The dashed (red) line shows 1-to-1. The color bar shows the count and applies to all subpanels.}
    \label{fig:ptype_spread_skill}
\end{figure}

Continuing the analysis, an essential step involves comparing the Brier score with various uncertainty estimates to assess how well these uncertainties align with true outcomes and the model's predictive performance. Figure~\ref{fig:ptype_spread_skill} displays 2D histograms that relate the computed Brier score to various uncertainties. The histograms are presented in four columns: aleatoric, epistemic, total, and entropy in (a), and DST $u$ in (b). Figure ~\ref{fig:ptype_spread_skill}(a) shows results from a deterministic MLP using Monte Carlo sampling, while \ref{fig:ptype_spread_skill}(b) is from an evidential MLP. 

Both models generally exhibit aleatoric uncertainty values that surpass the corresponding Brier scores, displaying an approximately sigmoidal correlation between skill and uncertainty. This relationship is more pronounced in the evidential model. Similarly, the epistemic uncertainty is usually outweighed by the Brier score for both models, with a notable drop in the Brier score at the highest epistemic uncertainties, particularly in the ensemble model. The evidential model generally predicts smaller or similar Brier scores for the same epistemic value.

The total uncertainty is comparable between the two approaches, with aleatoric uncertainty exerting a more significant influence than epistemic uncertainty. The ensemble model's overall uncertainty shape resembles that of the evidential model, though with more noise and a broader distribution along total aleatoric uncertainty.

In the ensemble model, the computed entropy closely mirrors the total uncertainty, primarily driven by aleatoric contributions. On the other hand, the shape of the DST $u$ histogram aligns more with epistemic uncertainty than with the total uncertainty. Both relationships exhibit a somewhat sigmoidal pattern rather than a linear one. When assuming a linear relationship, entropy becomes over-dispersed initially, while the evidential model remains under-dispersed. However, both eventually demonstrate over-dispersed behavior as $u$ increases, indicating the uncertainty surpasses the skill score. Additionally, both models show a slight over-dispersion for aleatoric uncertainty and under-dispersion for epistemic uncertainty.

\begin{figure}[t!]
    \centering
    \includegraphics[width=6 in]{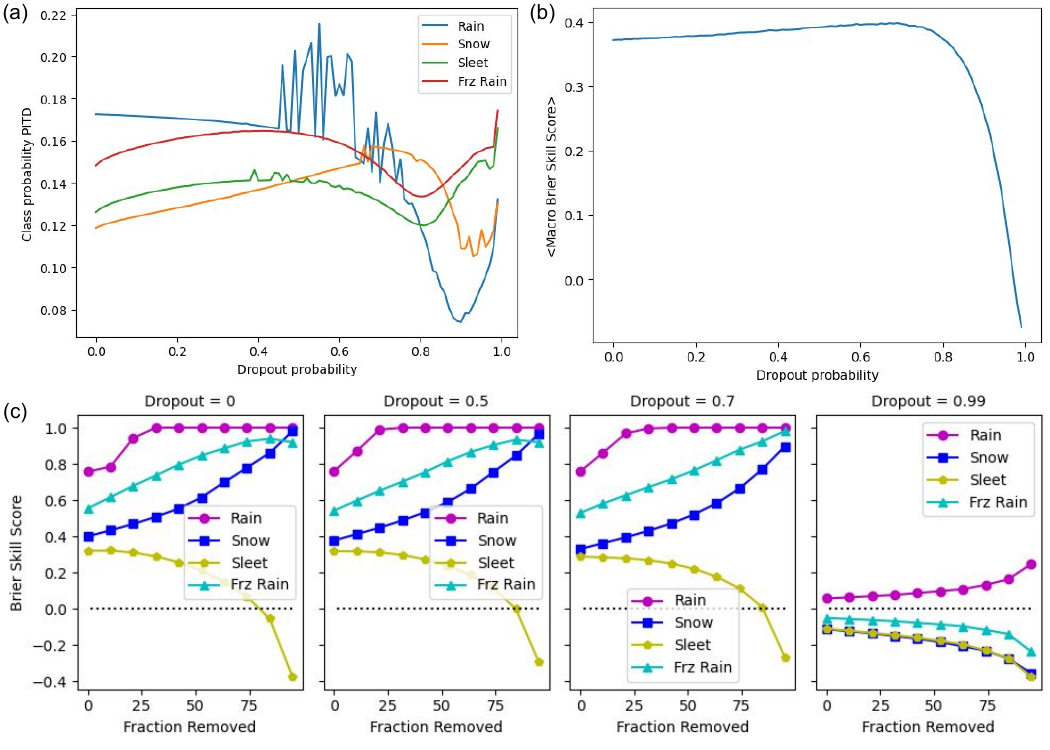}
    \caption{Dropout calibration analysis for the p-type deterministic model. (a) The PITD is calculated for the predicted class probabilities against varying dropout rates. (b) The macro Brier Skill Score is computed for each bin in the discard test and then averaged. The figure depicts the correlation between the skill score and the dropout rate. (c) Example discard-test outcomes obtained at different dropout rates.}
    \label{fig:s2}
\end{figure}

The dropout calibration analysis for the p-type deterministic model is shown in Figure~\ref{fig:s2}. In Figure~\ref{fig:s2}(a) PITD is computed for the predicted class probabilities against varying dropout rates, with the best dropout rate identified as the point when the curves reach their minimums, which are not the same value. In Figure~\ref{fig:s2}(b) the macro Brier Skill Score, which peaks just below 0.8, demonstrates a slightly lower optimal dropout rate when results are aggregated. Finally, in Figure~\ref{fig:s2}(c) The dropout rate shows limited sensitivity to the model's performance, and it becomes apparent that calibration results in lower performance on rain and snow but higher performance on sleet and freezing rain. These observations contribute to a better understanding of the model's behavior and its calibration with respect to different precipitation types.

\begin{figure}[t!]
    \centering
    \includegraphics[width=6 in]{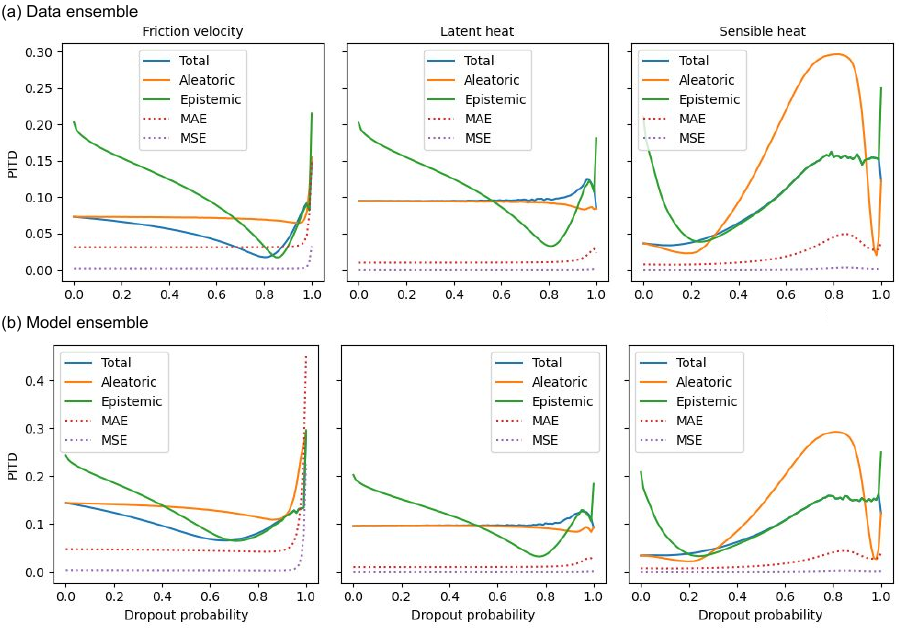}
    \caption{Comparison of the PITD against the dropout rate for the three tasks computed using aleatoric, epistemic, and total uncertainties. The MAE and MSE are also plotted in each sub-panel. (a) presents the results from the data ensemble, while (b) displays those from the deep model ensemble.}
    \label{fig:regression_calibration_supp}
\end{figure}

Figure~\ref{fig:regression_calibration_supp} illustrates the dropout calibration results for the three surface-layer flux models. In general, the two models exhibit similar calibration behavior, but there are notable distinctions for friction velocity. The deep ensemble shows PITD minimums at lower dropout rates compared to the data ensemble, suggesting better calibration in this specific case. Despite this difference, the overall shape of the calibration curves remains similar between the two models. For latent heat and sensible heat, the calibration results are more closely comparable, indicating consistent calibration performance for these flux models.

\begin{figure}[t!]
    \centering
    \includegraphics[width=6 in]{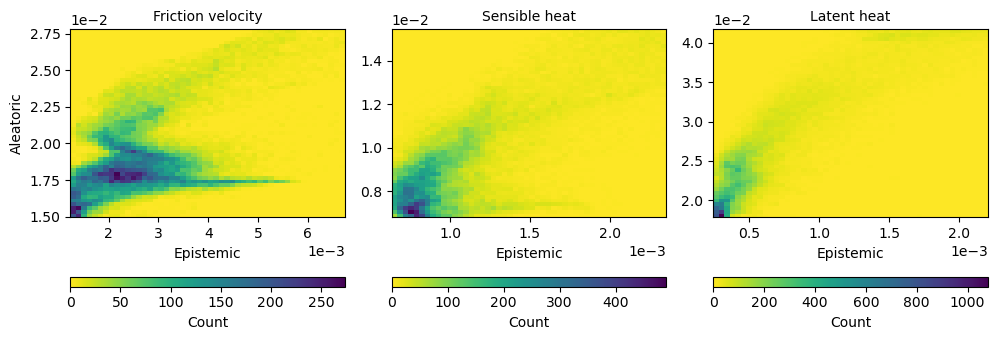}
    \caption{2D histograms illustrating the aleatoric and epistemic uncertainties for the models predicting friction velocity, sensible heat, and latent heat.}
    \label{fig:sl_ale_vs_epi}
\end{figure}

Figure~\ref{fig:sl_ale_vs_epi} shows 2D histograms comparing aleatoric and epistemic uncertainties for the three surface-layer flux quantities. The relationship between these uncertainties appears most complex for friction velocity, despite it being the simplest of the three processes.

\begin{figure}[t!]
    \centering
    \includegraphics[width=\columnwidth]{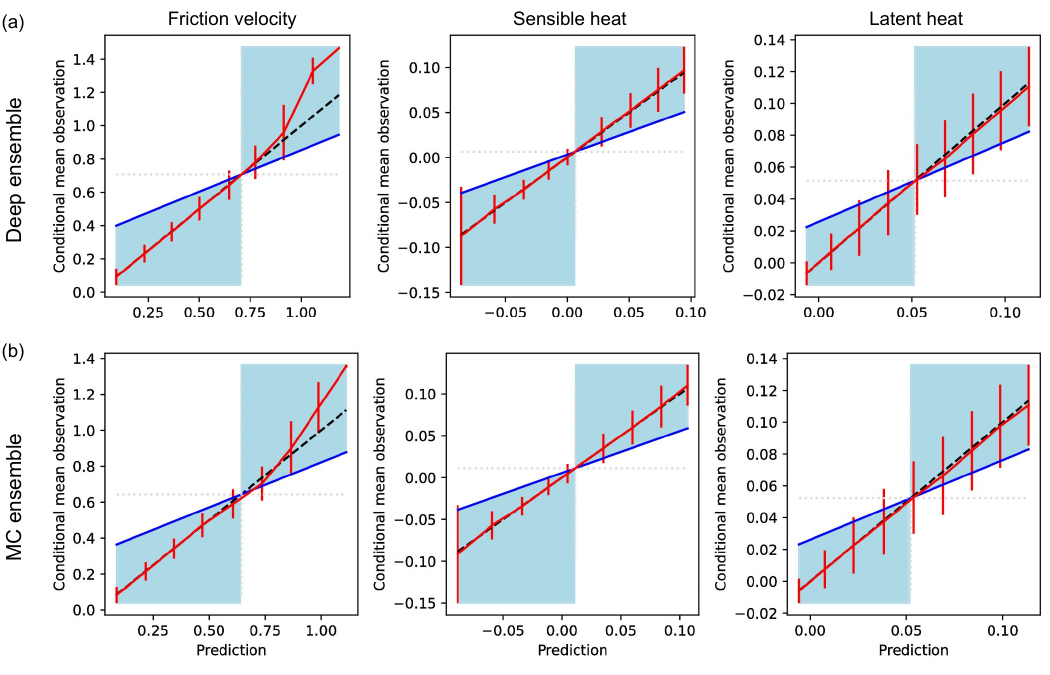}
    \caption{Attribution curves for (a) the deep ensemble utilizing the parametric Gaussian model and (b) the MC dropout ensemble generated using the best member from the deep ensemble. The reliability curve is represented by the red line. Each column displays the results for a specific model task. Similar to Figure~3 in the main text, the diagonal, horizontal, and vertical dashed lines in each subpanel indicate the 1-to-1 line, no-resolution line, and climatology line, respectively. The blue-shaded area signifies skill relative to climatology.}
    \label{fig:sl_attribute_supp}
\end{figure}

Figure~\ref{fig:sl_attribute_supp}(a) displays attribute diagrams computed for the deep ensemble consisting of 20 members. From this ensemble, the best model was selected, and Monte Carlo dropout was performed using the optimal dropout rate determined from Figure~\ref{fig:regression_calibration_supp}, creating an ensemble of size 100. The attribute diagram for the selected best model is shown in Figure~\ref{fig:sl_attribute_supp}(b).

\begin{figure}[t!]
    \centering
    \includegraphics[width=6 in]{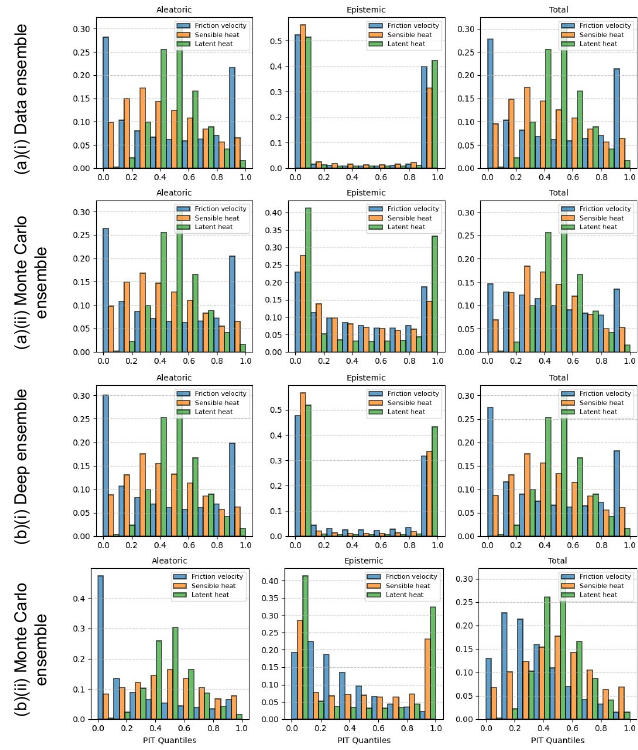}
    \caption{PIT histograms are depicted for the data and deep model ensembles in (a)(i) and (b)(i), respectively. Panels (a)(ii) and (b)(ii) display a Monte Carlo ensemble created using the 'best' model from the corresponding ensemble in (i), respectively.}
    \label{fig:sl_uq_all_pit}
\end{figure}
\newpage

Figure~\ref{fig:sl_uq_all_pit} presents the PIT histograms for each uncertainty type, computed using: the data ensemble, a Monte Carlo ensemble created using the best member from the data ensemble, a deep ensemble, and a Monte Carlo ensemble created using the best member from the deep ensemble. In most cases, the aleatoric uncertainty plays the dominant role in contributing to the total uncertainty, except for the Monte Carlo ensemble generated with the best data ensemble model where both quantities contribute to the total uncertainty. 

\begin{figure}[t!]
    \centering
    \includegraphics[width=6 in]{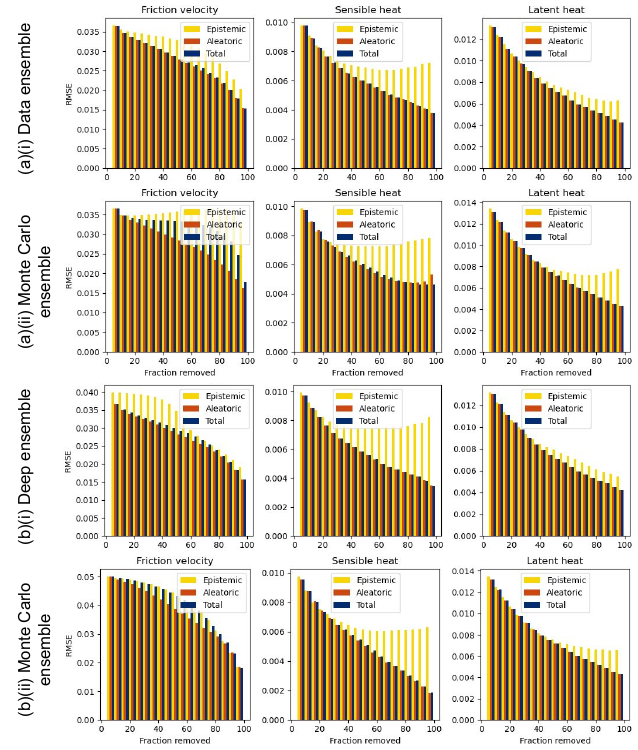}
    \caption{Discard-test histograms are presented for the data and deep model ensembles in (a)(i) and (b)(i), respectively. Panels (a)(ii) and (b)(ii) display a Monte Carlo ensemble created using the `best' model from the corresponding ensemble in (i), respectively.}
    \label{fig:sl_uq_all_df}
\end{figure}
\newpage

Complementary to Figure~\ref{fig:sl_uq_all_pit}, Figure~\ref{fig:sl_uq_all_df} presents the discard-test histograms for each uncertainty type, computed using: the data ensemble, a Monte Carlo ensemble created using the best member from the data ensemble, a deep ensemble, and a Monte Carlo ensemble created using the best member from the deep ensemble.

\begin{figure}[t!]
    \centering
    \includegraphics[width=4 in]{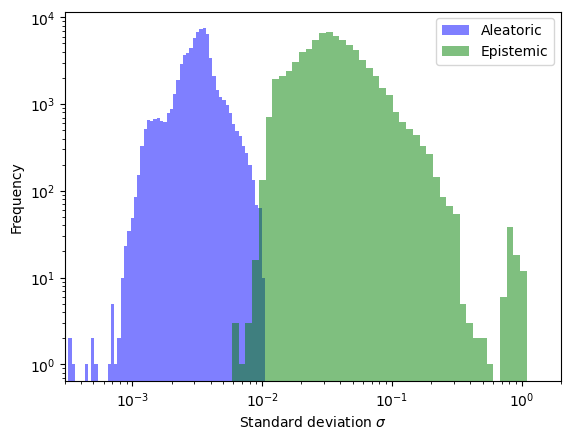}
    \caption{Distributions of aleatoric and epistemic quantities predicted by the friction velocity evidential model are depicted in histograms.}
    \label{fig:fv_histos}
\end{figure}

Figure~\ref{fig:fv_histos} plots histograms of the uncertainty predictions from the friction velocity model. The small peak observed on the far right may indicate potential outliers resulting from imperfect calibration of the model. 

\begin{figure}[t!]
    \centering
    \includegraphics[width=\columnwidth]{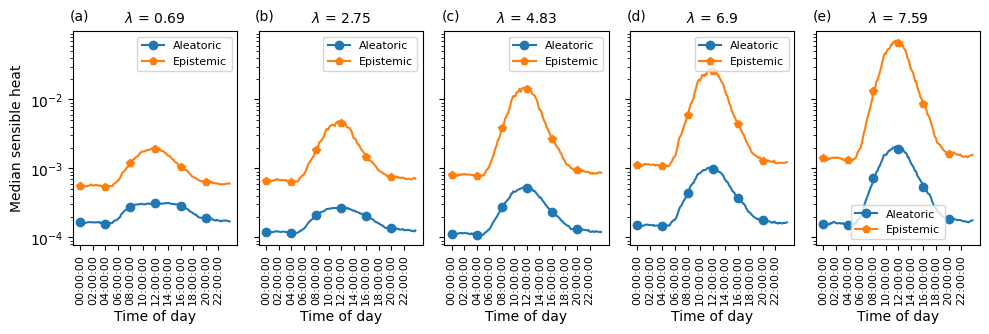}
    \caption{The median value of sensible heat in 10-minute increments is shown. The curves displayed in each panel represent the aleatoric, and epistemic quantities. All results are derived from the test data set.}
    \label{fig:sensible_heat_lambda}
\end{figure}

Figure~\ref{fig:sensible_heat_lambda} shows the dependence of the median aleatoric and epistemic versus time of the day. The title in each panel lists the value of the loss weight $\lambda$ in Equation~19 in the main text.

\acknowledgments
This material is based upon work supported by the National Science Foundation under Grant No. ICER-2019758 and by the National Center for Atmospheric Research, which is a major facility sponsored by the National Science Foundation under Cooperative Agreement No. 1852977. We would like to acknowledge high-performance computing support from Cheyenne and Casper \citep{Cheyenne} provided by NCAR's Computational and Information Systems Laboratory, sponsored by the National Science Foundation. MJM was supported by the University of Maryland Grand Challenges Program and the U.S. Department of Energy (DOE), Office of Science, RGMA component of the EESM Program under Award Number DE-SC0022070 and National Science Foundation IA 1947282.

\datastatement
All data sets used in this study are available at \url{https://zenodo.org/record/8368187}. The neural networks described here and the simulation code used to train and test the models are archived at \url{https://github.com/ai2es/miles-guess}.

\bibliographystyle{ametsocV6}
\bibliography{references}

\end{document}